\documentclass[sigconf]{acmart}

\usepackage{booktabs} %

\usepackage{times}  %
\usepackage{helvet}  %
\usepackage{courier}  %
\usepackage{url}  %
\usepackage{graphicx}  %

\usepackage{times}
\usepackage{latexsym}

\usepackage{url}

\usepackage{xcolor}
\usepackage{soul}
\usepackage[utf8]{inputenc}
\usepackage{caption}
\usepackage{tikz}
\usepackage{mathtools}
\usepackage{graphics}

\usepackage{natbib}
\setcitestyle{round}

\usepackage{amsmath}
\usepackage{amsfonts}
\usepackage{multirow}
\usepackage{flushend}
\usepackage{makecell}
\usepackage{subcaption}
\usepackage{soul}

\newcommand{\myrightarrow}[1]{\mathrel{\raisebox{-2pt}{$\xrightarrow{\mathrm{#1}}$}}}
\newcommand{\myleftarrow}[1]{\mathrel{\raisebox{-2pt}{$\xleftarrow{\mathrm{#1}}$}}}
\newcommand{\mydoublearrow}[1]{\mathrel{\raisebox{-2pt}{$\xleftrightarrow{\mathrm{#1}}$}}}
\renewcommand{\vec}[1]{\mathbf{#1}}

\DeclareMathOperator*{\argmax}{arg\,max}

\graphicspath{{./figures/}}
\usepackage{hhline}

\usepackage{xcolor}
\renewcommand{\textrightarrow}{$\rightarrow$}

\newif\ifextended
\extendedtrue %

\copyrightyear{2019}
\acmYear{2019}
\setcopyright{iw3c2w3}
\acmConference[WWW '19]{Proceedings of the 2019 World Wide Web Conference}{May 13--17, 2019}{San Francisco, CA, USA}
\acmBooktitle{Proceedings of the 2019 World Wide Web Conference (WWW'19), May 13--17, 2019, San Francisco, CA, USA}
\acmPrice{}
\acmDOI{10.1145/3308558.3313486}
\acmISBN{978-1-4503-6674-8/19/05}

\usepackage{balance}

\begin{document}

\title{Predicting ConceptNet Path Quality\\Using Crowdsourced Assessments of
Naturalness}

\begin{abstract}
In many applications, it is important to characterize the way in which two concepts are semantically related. Knowledge graphs such as ConceptNet provide a rich source of information for such characterizations by encoding relations between concepts as edges in a graph. When two concepts are not directly connected by an edge, their relationship can still be described in terms of the paths that connect them.
Unfortunately, many of these paths are uninformative and noisy, which means that the success of applications that use such path features crucially relies on their ability to select high-quality paths. In existing applications, this path selection process is based on relatively simple heuristics.
In this paper we instead propose to learn to predict path quality from crowdsourced human assessments. Since we are interested in a generic task-independent notion of quality, we simply ask human participants to rank paths according to their subjective assessment of the paths' \emph{naturalness}, without attempting to define naturalness or steering the participants towards particular indicators of quality. We show that a neural network model trained on these assessments is able to predict human judgments on unseen paths with near optimal performance. Most notably, we find that the resulting path selection method is substantially better than the current heuristic approaches at identifying meaningful paths.

\end{abstract}




\author{Yilun Zhou}
\affiliation{\institution{MIT CSAIL}}
\email{yilun@mit.edu}

\author{Steven Schockaert}
\affiliation{\institution{Cardiff University}}
\email{SchockaertS1@cardiff.ac.uk}

\author{Julie A. Shah}
\affiliation{\institution{MIT CSAIL}}
\email{julie\_a\_shah@csail.mit.edu}

\maketitle

\section{Introduction}

Many applications require information about the semantic relation between two (or more) words, concepts, or entities. For example, a recommender system needs to recommend items related to a user's browsing history, a task allocation agent needs to match people's experiences and skills to a pool of problems to maximize problem-solving efficiency, and
a household robot given the instruction to ``wash the plates'' needs to infer that a dishwasher could be used.
Open-domain knowledge graphs such as ConceptNet \citep{speer2017conceptnet} allow us to characterize such semantic relationships in the form of relational paths. Compared to the use of word embeddings \citep{mikolov2013distributed} for characterizing relatedness, knowledge graphs have the potential advantage of producing easier-to-understand characterizations, and they can capture relationships that go beyond what is encoded in standard word embeddings \citep{xu2014rc}.

Typical knowledge graphs (KGs), such as DBpedia and WikiData, are concerned with named entities and their relations (e.g. ``Abraham Lincoln'' is one of ``United States Presidents'').
In this paper, however, we are concerned with capturing semantic relationships between common nouns or concepts, which requires a commonsense KG such as ConceptNet. Despite its indisputable value, however, effectively using ConceptNet in applications comes with a unique set of challenges. The knowledge captured in ConceptNet is, by design, often informal, subjective and vague. Due to the fact that ConceptNet partly consists of unverified crowdsourced assertions, it is also noisier than many other KGs.
 Furthermore, many commonsense assertions are true only under some circumstances and to some extent. For example, ConceptNet encodes that ``popcorn'' is required for ``watching a movie'', which captures the useful commonsense knowledge that eating popcorn is associated with watching a movie, but the statement that popcorn is \emph{required} is nonetheless false. In addition, ConceptNet only disambiguates concepts to a coarse part-of-speech level (e.g. noun meaning vs verb meaning of the word ``watch''). Finally, a lot of concepts are linked by the generic ``RelatedTo'' relation, which covers relationships as diverse as collocations (``Tire $\mydoublearrow{RelatedTo}$ Spare''), hypernyms (``Tire $\mydoublearrow{RelatedTo}$ All Seasons Tire''), co-meronyms (``Tire $\mydoublearrow{RelatedTo}$ Exhaust''), homonyms (``Tire $\mydoublearrow{RelatedTo}$ Tier''), and very loosely related terms (``Tire $\mydoublearrow{RelatedTo}$ Clothes'').

Because of those challenges, few authors use relational paths from ConceptNet in applications. In particular, while several authors have found the knowledge encoded in ConceptNet to be highly valuable, they typically restrict themselves to using relationships that are directly encoded as an edge. For example, \citet{speer2017conceptnet} showed that ConceptNet can be used to improve word embeddings by incorporating the intuition that words which are linked by an edge in ConceptNet should be represented by similar vectors. We believe that this common restriction to direct edges is due to the lack of sufficiently accurate methods for filtering nonsensical paths, or conversely, for identifying the most natural paths. This path selection problem is the focus of our paper.

In existing work \citep{gardner2014incorporating,lin2015modeling}, the problem of selecting high-quality ConceptNet paths has been addressed in a heuristic way (see Section \ref{secExperiments}). While some intuitions about high-quality paths can easily be formulated (e.g.\ shorter paths tend to be more informative than longer paths, the nodes occurring in natural paths tend to have similar word vector representations), such heuristic methods still fail to filter out many nonsensical paths, and conversely sometimes erroneously filter out highly valuable paths. For example, the best path between ``lead'' and ``poison'' found by one of the standard heuristics (pairwise baseline, see Section \ref{pred_perf}) is ``Lead $\mydoublearrow{Synonym}$ Take $\mydoublearrow{DistinctFrom}$ Give $\mydoublearrow{RelatedTo}$ Poison'', which uses the verb meaning of ``lead'', despite the  fact that the noun meaning (i.e.\ the poisonous chemical element Pb) is more relevant in this case.

Rather than trying to construct increasingly intricate heuristics, we propose to learn to predict path quality using a neural network model. To this end, we rely on crowdsourced assessments about the \emph{naturalness} of ConceptNet paths. Specifically, to collect training data, human annotators were asked to choose the more natural path among a pair of paths, without any guidance on how to interpret naturalness. This notion of naturalness was chosen because it is intuitively easy to understand for crowdsourcers (as opposed to e.g.\ terms such as ``semantic coherence'' or ``predictive value''). The term is also deliberately vague, as we do not want to steer annotators towards particular types of features. The resulting pairwise judgments are then used to train a neural network to predict a latent naturalness score, which can be used for path ranking or path selection.

The main contribution of our work is to show that people's intuitive understanding of naturalness is sufficiently coherent to be used as training data for a data-driven path selection method. To this end, we train a simple neural network model to predict latent naturalness scores that are predictive of human pairwise judgments. We find that
our model can predict such judgments with a performance that is close to the optimum suggested by the inter-annotator agreement (Sec \ref{multiassign_sec}). For example, the best path found by our model for the above example is ``Lead $\myrightarrow{HasProperty}$ Toxic $\mydoublearrow{RelatedTo}$ Lethal $\mydoublearrow{RelatedTo}$ Poison''. We also show, for a number of different evaluation tasks, that our method allows us to select semantically more meaningful paths than the previously proposed heuristics.

\section{Related Work}
\textbf{Knowledge graphs:} KGs explicitly encode relationships between different entities as subject-predicate-object triples. Such triples can be seen as defining a graph, where the subject and object entities refer to nodes of the graph and the predicates correspond to edge labels. KGs are typically constructed by a domain expert, via crowdsourcing, or by extracting assertions from a natural language corpus \cite{carlson2010toward}. In this work, we used ConceptNet \citep{liu2004conceptnet}, one of the most comprehensive commonsense knowledge graphs with 46 relation types, nearly 1 million concepts (words and phrases), and nearly 3 million edges (i.e.\ triples). ConceptNet partly obtained through crowdsourcing, but also incorporates external sources such as OpenCyc \citep{lenat1995cyc}, WordNet \citep{miller1990introduction}, Verbosity \citep{von2006verbosity}, DBPedia \citep{auer2007dbpedia}, and Wiktionary\footnote{\url{https://www.wiktionary.org/}}.

\smallskip
\noindent
\textbf{Knowledge base construction by crowdsourcing:}
While the use of crowdsourcing for constructing knowledge bases has already been well studied, most existing approaches focus on knowledge acquisition. This can involve, for instance, direct collaborative editing of a knowledge base \citep{bollacker2008freebase, vrandevcic2014wikidata} or indirect construction of a knowledge base through the use of an interactive game \citep{von2006verbosity}.
 In contrast to these works, our focus is on using crowdsourcing for learning how to detect noisy paths within an existing knowledge graph.

An important challenge with crowdsourcing is the fact that there will inevitably be disagreements within the collected data. One framework \citep{aroyo2014three} suggests that this can be due to (i) the inclusion of low-quality participants, (ii) an ambiguous interpretation of the input data and (iii) an ambiguous definition of the labels that crowdsourcers are required to provide. In our work, we mitigated the first issue with a quality control mechanism (Section \ref{datacollection}). To address the remaining two issues, we will rely on a probabilistic generative model to interpret the provided ratings.

\smallskip
\noindent \textbf{Word embeddings:}
Word embedding methods \cite{mikolov2013distributed,pennington2014glove} represent each word in a relatively low-dimensional space of typically around 300 dimensions, which is estimated from word co-occurrence statistics. With most training procedures, vector differences represent abstract relations in the resulting embedding space: for example, $\vec v_{\mathrm{king}}-\vec v_{\mathrm{queen}}\approx\vec v_{\mathrm{man}}-\vec v_{\mathrm{woman}}$, in which $\vec v_{\mathrm{word}}$ represents the embedding of a given word.

Both KGs and word embeddings capture aspects of meaning, and can thus to some extent be seen as alternatives. One advantage of word embeddings is that they capture types of knowledge which are difficult to encode symbolically. For example, the cosine similarity between word vectors tends to correlate very strongly with human perceptions of word similarity. On the other hand, although related words are often close to each other in the embedding space, it is difficult to ``reverse-engineer'' and describe the specific relation from the vector difference \citep{DBLP:conf/coling/BouraouiJS18}. Moreover, when multiple relations exist between two concepts (e.g. ``France $\mydoublearrow{BorderWith}$ Germany'' and ``France $\mydoublearrow{AllyWith}$ Germany''), the vector difference can only reflect an aggregate. Apart from issues which arise because of the use of vectors, there are also some problems that are more generally related to the use of co-occurrence statistics for representing meaning. As a simple illustration of such problems, a system \cite{krebs2018semeval} incorrectly assumes that garlic has wings, because of the prevalence of phrases such as ``garlic chicken wings''.

Given the complementary nature of the KGs and word embeddings, it is perhaps not surprising that several studies have found it to be beneficial to integrate these two types of resources. Some studies (e.g., \citep{speer2017conceptnet, faruqui2014retrofitting, xu2014rc}) reported that knowledge graphs can be used to improve embeddings to achieve better results on various benchmark tasks such as synonym selection and analogy question solving. Conversely, embeddings can also be incorporated into methods that rely on paths in knowledge graphs, for example to cluster such paths \cite{gardner2014incorporating}, or as features for a path selection method, as in our work.

\smallskip
\noindent\textbf{Path Features in Applications:}
Several prior works have described systems that incorporated knowledge graph paths as features. For example, \citet{boteanu15solving} used ConceptNet to solve analogy questions (e.g., ``dog is to animal as banana is to fruit'') by comparing similarities between paths.
\citet{lin2015modeling} proposed models for learning the embeddings of paths and showed that they performed better on problems such as relation prediction compared to embeddings computed by previous methods.
\citet{guu2015traversing} proposed to answer compositional queries (e.g., ``Q: What is the population of the capital of Russia?'' ``A: Russia $\myrightarrow{Capital}$ Moscow $\myrightarrow{Population}$ 12.2 million'') by traversing the knowledge graph in \em vector space\rm, representing traversal as a series of matrix-vector multiplications. \citet{das2017chains} proposed a recurrent neural network (RNN) model with attention mechanism to reason over entities and relations in a knowledge graph, and observed improved performance in path query answering upon that by \citet{guu2015traversing}.

Among many applications, knowledge graph completion (i.e., inferring missing relations from existing paths) has been a popular target.
\citet{lao2010relational} used path features for knowledge graph completion in the path ranking algorithm (PRA). %
\citet{gardner2014incorporating} incorporated embeddings of the \em relations\rm, allowing the system to recognize semantically similar relations, such as ``(river) runs through (city)'' and ``(river) flows through (city).''
\citet{neelakantan2015compositional} built upon the same PRA idea, but used an RNN to model paths.
\citet{toutanova2016compositional} proposed a dynamic programming algorithm that can incorporate both edge and vertex features and reported better performance on knowledge graph completion.
While all of the above-mentioned knowledge graph completion works use some version of PRA to generate paths, \citet{xiong2017deeppath} trained a reinforcement learning agent for path generation that rewards accuracy, efficiency and diversity, and demonstrated better performance than PRA-generated paths.

\smallskip
\noindent\textbf{Path Selection:}
The number of paths between two nodes in a knowledge graph tends to grow exponentially with path length. In addition, while each type of path can be viewed as encoding a kind of semantic relationship, for many paths this relationship is difficult to interpret. Thus, all of the works described above mitigate the scalability and/or quality problem via heuristics. To limit the number of paths, \citet{boteanu15solving} stopped the path search after a pre-defined number of nodes are explored. \citet{lao2010relational} and \citet{guu2015traversing} limited the maximum path length. To obtain natural paths, \citet{gardner2014incorporating} favored paths that contain words with higher embedding similarity. \citet{lin2015modeling} calculated the quality of paths using a heuristic based on vertex degree and network flow.

We note that while ConceptNet has been proven useful, most works only use direct edges, with one work \citep{boteanu15solving} being a notable exception. For other types of knowledge graphs, it was already shown that incorporating longer relational paths can be highly beneficial, and there is no reason to assume that this situation would be any different for ConceptNet. For instance, while there is an intuitively obvious semantic relationship between the words ``beach'' and ``sun'', there is no direct edge connecting these words in ConceptNet. However, this relationship can be uncovered by observing that ConceptNet contains the path ``Beach $\mydoublearrow{RelatedTo}$ Sunbathing $\mydoublearrow{RelatedTo}$ Sun''.
Nevertheless, to enable more effective use of such relational paths from ConceptNet, we believe that more work is needed on how to avoid nonsensical paths, which is the focus of this paper.

\section{Method}

\subsection{Problem Formulation}
\label{formulation}
Our goal in this work is to train a classifier for predicting path quality based on crowdsourced information. Specifically, we ask human annotators to assess the \emph{naturalness} of ConceptNet paths. Rather than trying to provide guidelines on how naturalness should be understood, we simply rely on annotators' intuitive understanding of this notion. This has several advantages, including the fact that we do not steer annotators towards particular indicators and the fact that this makes the annotation task much easier. Moreover, given that we are interested in a task-independent form of path quality (i.e., our focus is on eliminating non-sensical and uninformative paths, rather than on selecting the best paths for a particular application), it is not clear what further guidance would be meaningful for annotators. As we will discuss in more detail in Section \ref{multiassign_sec}, despite the lack of guidance on how naturalness should be understood, we observed a high inter-annotator agreement in practice.

Clearly, however, asking participants to provide ratings on an absolute scale is problematic: people have varying thresholds for naturalness, and these thresholds may also change over time. After observing a large number of unnatural paths, people may lower their thresholds to be more willing to accept a path as natural, and vice versa.
Therefore, we instead asked participants about pairwise comparisons: determining which of two given paths is more natural. In this setting, if one path is more natural than the other, two raters would provide the same answer even if they maintained different absolute naturalness thresholds, or if they shifted their underlying absolute threshold unconsciously over time.

If the two paths are equally (un)natural, the selected answer would be more or less arbitrary, and even the same annotator might not consistently give the same answer when presented with the same pair more than once. Therefore, in our model, we assume that answers to these pairwise comparison questions are probabilistically generated from a latent naturalness score. Specifically, each path is assumed to have a latent naturalness score $m$, such that for a pair paths with scores $m_1$ and $m_2$, the observed answer is drawn from a Bernoulli distribution, with probability $e^{m_1}/(e^{m_1}+e^{m_2})$ that the first path is selected. The problem we consider is to learn a model that can predict the naturalness score $m$ of a given path, using crowdsourced answers of pairwise comparisons as supervision signal.

\subsection{Model}
In the big picture, our model consists of an encoder and a pairwise predictor. The encoder is a long short-term memory (LSTM) network \citep{hochreiter1997long} that transforms a path into a code vector. The predictor then takes the path codes and computes the probability that one path is more natural than the other. Each path is represented by an alternating sequence of vertex and edge representations, $(\allowbreak v_1, \allowbreak e_1, \allowbreak v_2, \allowbreak e_2, ..., \allowbreak e_{n-1}, \allowbreak v_n)$. Each vertex representation is a list of $n_v$ features, $v_i=(\allowbreak \vec v_{i,1}, \allowbreak \vec v_{i,2}, \allowbreak ..., \allowbreak \vec v_{i,n_v})$, with $j$-th feature being a $d_j^v$-dimensional vector. The edge representations are structured similarly. The features which we use are summarized in Section \ref{features}.

\begin{figure}[!htb]
  \centering
  \begin{subfigure}{2.5in}
    \includegraphics[width=2.5in]{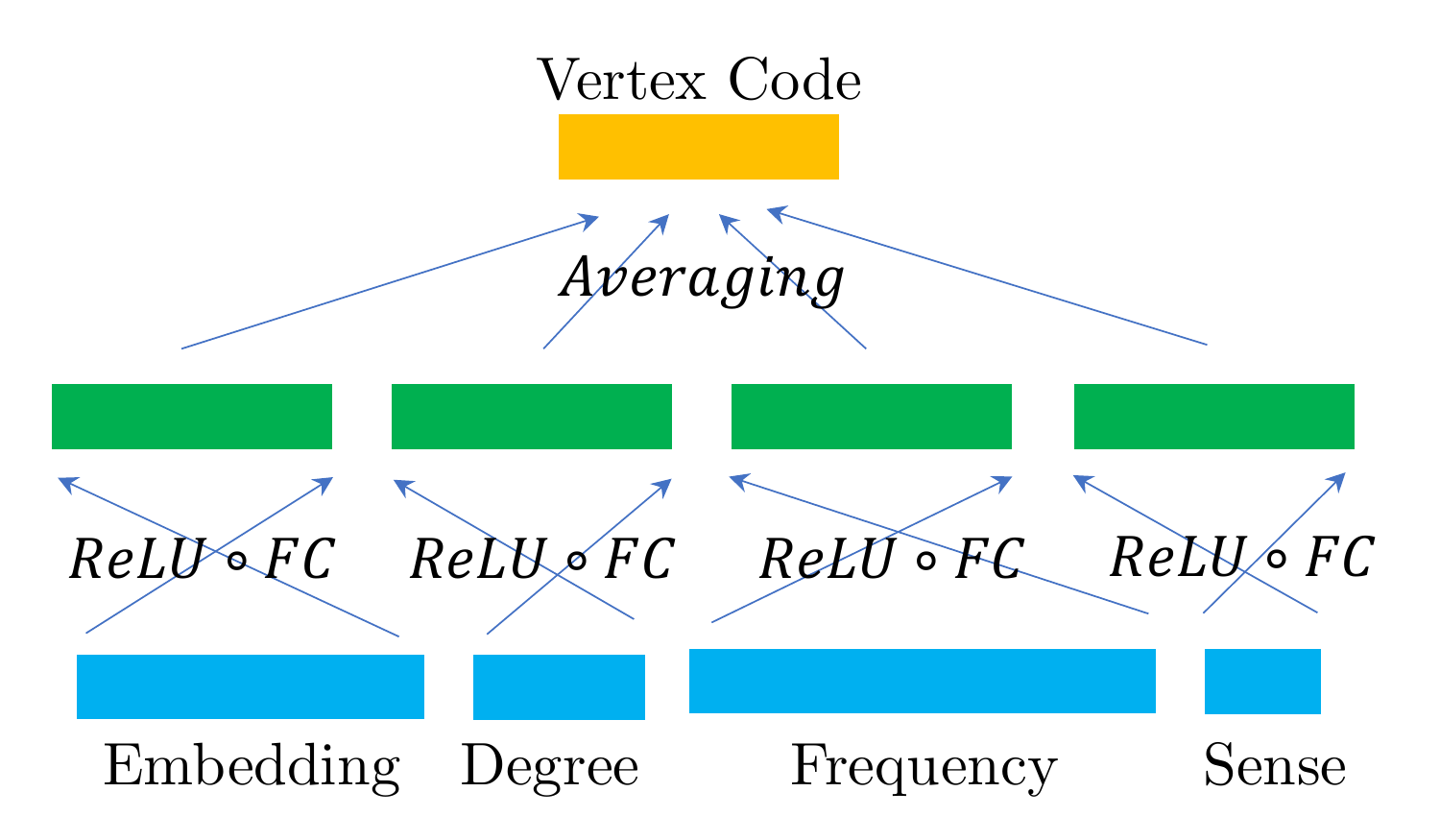}
    \caption{The encoding of a vertex. }
    \label{v_encoding}
  \end{subfigure}
  \begin{subfigure}{2.5in}
      \includegraphics[width=2.5in]{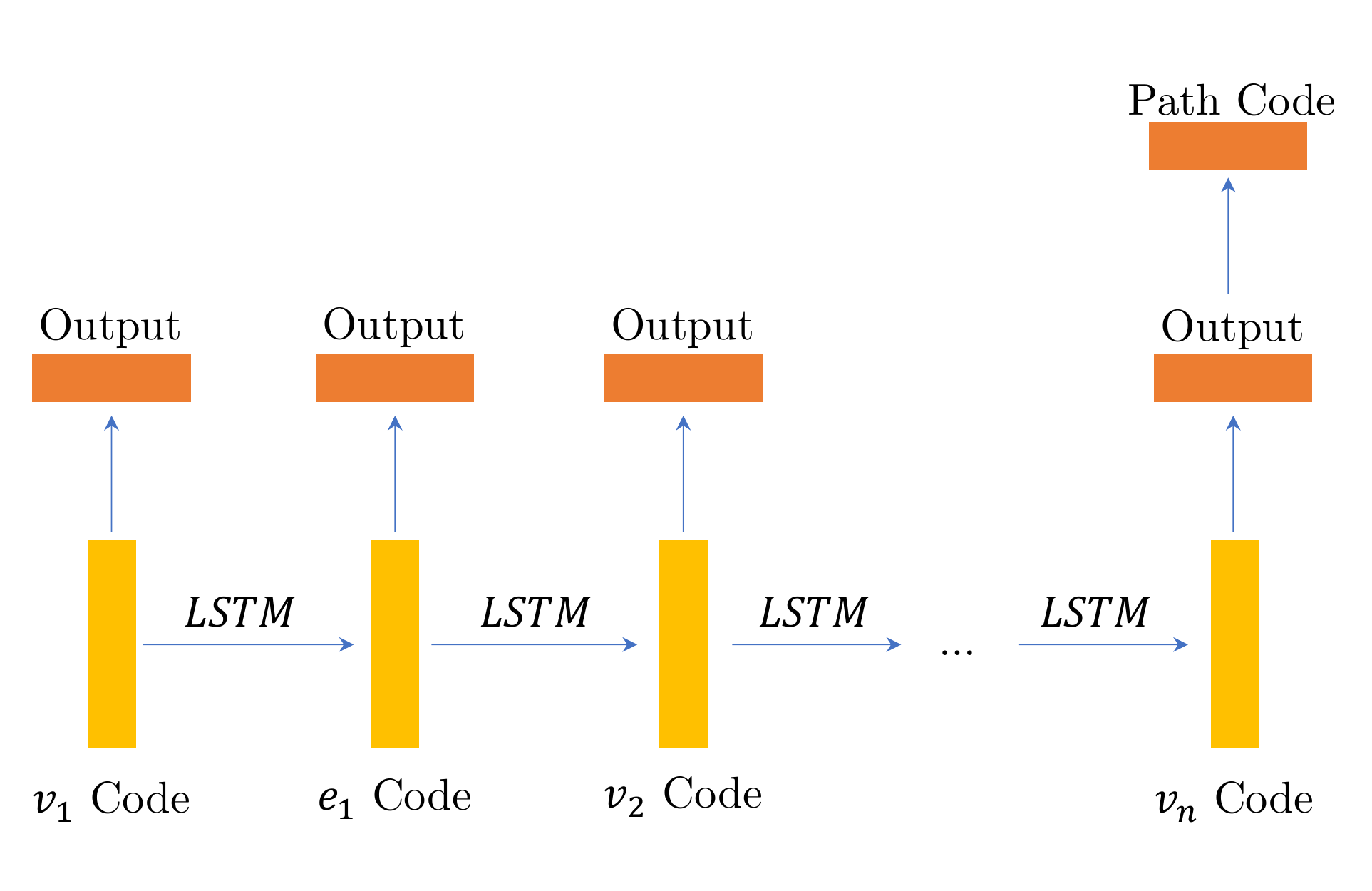}
      \caption{The encoding of the path. } \label{chain_encoding}
  \end{subfigure}
  \begin{subfigure}{2.5in}
  \includegraphics[width=2.5in]{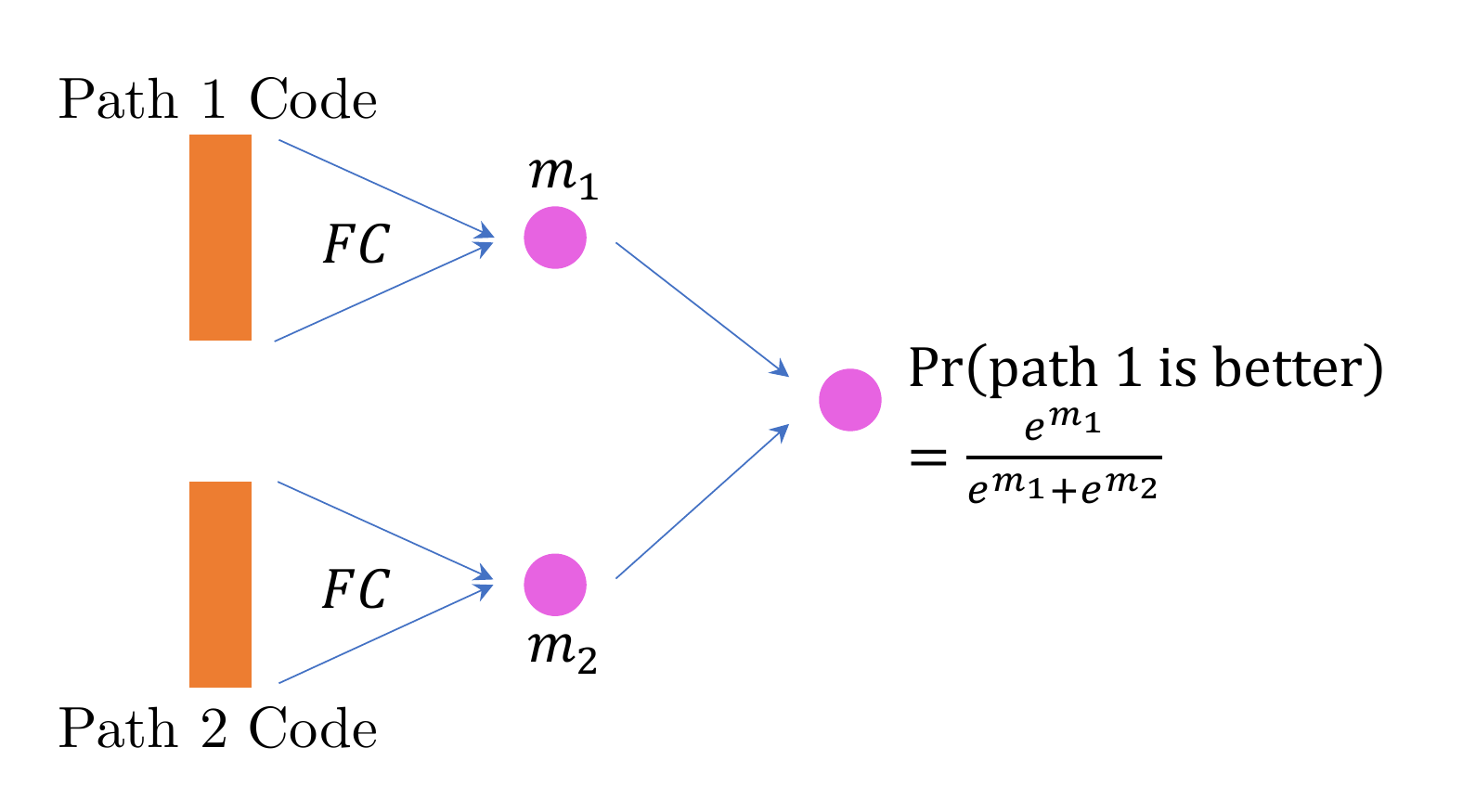}
  \caption{The predictor model. }\label{predictor}
  \end{subfigure}
  \caption{LSTM model architecture. Blue (bottom) cells represent raw features (of variable lengths). Green (middle) cells represent transformed features (of length $l_f$). Orange (top) cells represent the final code for the vertex (of length $l_f$). The encoding of an edge is computed in a similar fashion. The codes (yellow) for vertices and edges are successively processed by an LSTM network. The last state $\vec h_{2n-1}$ (orange) is used as the code for the entire path. The path codes (orange) first pass separately through fully-connected layers (with shared weight) with an output dimension of 1. A softmax layer then calculates the probability that path 1 is more natural than path 2. }\label{architecture}
\end{figure}

A standard LSTM architecture can only process sequential data of the same dimension. Thus, vertex and edge representations must first pass through an encoder. Figure \ref{v_encoding} depicts how a neural network encodes a vertex into a vertex code of pre-defined length $l_f$. Each feature (such as the word embedding and absolute frequency in some large corpus) is transformed through a rectified linear unit (ReLU)-activated fully-connected (FC) layer to a vector of length $l_f$. The overall code is the average of the vectors. The encoding of an edge to the same length $l_f$ is conducted in a similar fashion.
Figure \ref{chain_encoding} depicts the way in which the path code is calculated. After the codes for each vertex and edge are computed, they are sequentially aggregated by an LSTM network. The last state $\vec h_{2n-1}$ is interpreted as the code of the path.
As shown in Figure \ref{predictor}, the predictor then outputs the pairwise comparison result. It first transforms each path code to a score ($m_1$ and $m_2$) using linear layers with shared weight. The softmax function is then applied to the two scores to calculate a probability. While the network predicts pairwise comparisons, it is the score $m_i$ produced by the network -- or, at least, the ordering induced by the scores -- that matters for applications.

The problem we consider falls under the broad umbrella of learning to rank, for which many algorithms have already been proposed. We use an LSTM encoder because it elegantly deals with variable length input, while most other methods can only take fixed length input representations.

\subsection{Training}
\label{training}
We used the negative-log-likelihood (NLL) loss as our objective function, along with the Adam optimization algorithm \citep{kingma2014adam}. Although the neural network model seems like a natural choice for the data generation process described in Section \ref{formulation}, here we show that the predicted scores $m$ indeed converge to the latent scores from which the pairwise comparison results were generated, up to a constant difference.

Consider two paths with scores $m_1$ and $m_2$. According to our generative model, the first path is selected with probability $p_{m_1m_2} = e^{m_1}/(e^{m_1}+e^{m_2})$.  Note that from $e^{m_1}/(e^{m_1}+e^{m_2})=e^{m_1'}/(e^{m_1'}+e^{m_2'})$ it follows that $m_1-m_1'=m_2-m_2'$ by simple algebra. This means that if our system can correctly predict the probabilities $p_{m_1m_2}$, for any pair of paths, then it must be the case that there is a fixed constant $c$ such that for each path with naturalness score $m$, the predicted naturalness score $m'$ is such that $m=m'+c$.

It remains to be shown that the NLL objective will indeed lead our neural network model to predict the correct probabilities. Specifically, assume that observations $\{y_i\allowbreak\sim \allowbreak\mathrm{Bernoulli}\allowbreak(q_{\theta}(\vec x_i))\}_{i=1}^n$ are given, where $\vec x_i$ is the feature representation of a given pair of paths and $q_{\theta}(\vec x_i)$ is the corresponding output of our neural network model when the parameters are set as $\theta$. We show that minimizing the NLL objective will result in parameters $\phi$ such that $q_{\theta}=q_{\phi}$.
For any fixed $\vec x_i$, given a set of observations $\{(\vec x_i, y_{i,1}), ..., (\vec x_i, y_{i,n_i})\}$, we define the expected data likelihood with respect to $q_i$ to be as follows:
\begin{align*}
  \mathbb E_{q_i}\left[\Pr(y_{i,1:n_i}|\vec x_i)\right] = \mathbb E\left[\prod_{j=1}^{n_i}\left[y_{i,j}q_i+(1-y_{i,j})(1-q_i)\right]\right].
\end{align*}
\noindent When $y_{i,1:n_i}$ is generated by Bernoulli $(q_{\theta}(\vec x_i))$, it can be shown that $\argmax_{q_i} \allowbreak\mathbb E_{q_i}\left[\Pr(y_{i,1:n_i}|\vec x)\right]\allowbreak=\allowbreak q_{\theta}(\vec x_i)$. For the entire dataset $\{(\vec x_1, y_1), \allowbreak.., \allowbreak(\vec x_n, y_n)\}$, since the $y_i$'s are independent of each other given $\vec x_i$, the expected joint likelihood for the entire dataset
\begin{align*}
\mathbb E_{q_{\phi}}\left[\Pr(y_{1:n}|\vec x_{1:n})\right] =  \mathbb E\left[\prod_{j=1}^{n}\left[y_{j}q_{\phi}(\vec x_j)+(1-y_{j})(1-q_{\phi}(\vec x_j))\right]\right]
\end{align*}
 is then maximized for $q_{\phi}(\vec x)=q_{\theta}(\vec x)$ for all $\vec x$. Therefore, we can simply minimize NLL (i.e. $-\log\left[\Pr(y_{1:n}|\vec x_{1:n})\right]$).

Note in particular that the network will converge to optimal predictions even with only one answer for each pair, which we can exploit to maximize the diversity of the crowdsourced annotations: by only collecting a single annotation for each pair of paths, a broader range of paths can be assessed (with a fixed budget) than with multiple annotations per pair. While each individual answer will to some extent be arbitrary (e.g., the fact that path 1 was selected could mean that path 1 is more natural or that both paths are approximately equally natural and that path 1 was selected by chance), because the network is trained on many different pairs, it will still learn to differentiate between clear-cut cases and borderline cases.

\subsection{Features}
\label{features}
In this section, we describe the features we used to encode vertices and edges. The number in parenthesis after each feature name indicates the dimension of that feature. %

\smallskip
\noindent\bf Vertex embedding (300) \rm This feature is taken directly from the 300-dimensional GloVe \citep{pennington2014glove} embedding, pre-trained on the Common Crawl\footnote{\url{http://commoncrawl.org/}} dataset with 840 billion tokens. In some experiments, we used principal component analysis (PCA) to first reduce the dimensionality of this feature before inputting it into the neural network.

\noindent\bf Vertex frequency (1) \rm This feature is a scalar representing the frequency of the given word. We estimated the (unnormalized) frequency using Zipf's law \citep{zipf1935psycho} from word occurrence ranks, which can be derived from the pre-trained GloVe embedding since it orders words by frequency. For example, the word ``science'' is ranked 717th, and is thus assigned a frequency of 1/717=1.39e-3.

\noindent\bf Vertex degree (1) \rm This feature is the number of neighbors (in both directions) of the vertex in the graph, representing how well connected the word is within ConceptNet.

\noindent\bf Vertex sense score (1) \rm This feature is a number between 0 and 1 representing how consistent is the overall meaning of the vertex compared to its neighbors on the left and right. For example, the sense score for the word ``book'' in the path ``Knowledge $\myleftarrow{HasA}$ Book $\mydoublearrow{RelatedTo}$ Restaurant'' would be quite low because it is used in two different senses
. On the other hand, the sense score for ``book'' in ``Knowledge $\myleftarrow{HasA}$ Book $\mydoublearrow{RelatedTo}$ Paper'' would be high. We believe that a lower sense score would result in the path being considered less natural. In calculating this feature, we borrow and extend the idea of ``word sense profile'' proposed by \citep{chen2011combining}.
\ifextended
We leave the details to the appendix at the end of the paper.
\else
The details of the sense score calculation can be found in the appendix of the extended version available at \url{https://arxiv.org/XXXXXX}.
\fi

\noindent\bf Edge ends similarity (1) \rm For an edge connecting two vertices $(s, t)$, this feature is the cosine similarity of the embeddings of $s$ and $t$.

\noindent\bf Edge direction (3) \rm This feature is a one-hot vector that represents if the edge is forward (``Dog $\myrightarrow{IsA}$ Animal''), backward (``Wheel $\myleftarrow{HasA}$ Car''), or bidirectional (``Big $\mydoublearrow{Antonym}$ Small'').

\noindent\bf Edge relation (46) \rm This feature is a one-hot vector that represents the relation type of the edge.

\noindent\bf Edge provenance (6) \rm In ConceptNet, each edge is derived from at least one source. In the case of multiple sources, the weight is the sum of the weights across all sources. However, weights across sources are not directly comparable; for example, nearly all edges from WordNet have a weight of 2.0, while those from Wiktionary have a weight of 1.0. This does not necessarily mean that WordNet is more reliable than Wiktionary. Thus, we encode the provenance of an edge using a 6-dimensional vector, with one component for each possible source: WordNet, DBpedia, Verbosity, Wiktionary, OpenCyc, and Open Mind Common Sense \citep{singh2002open}. When an edge has only one source, its component in the vector has a value equal to the weight, and all other entries are 0. When an edge has multiple associated sources, we split the weight across the sources using the most common weight of each source (e.g., 2.0 for WordNet).

\noindent\bf Edge sense score (1) \rm Similar to the vertex sense score, this feature is also a number between 0 and 1 characterizing the consistency of meaning, but at an edge level.
\ifextended
The details can again be found in the appendix.
\else
Again, the details are included in the extended version.
\fi

\section{Experiments}\label{secExperiments}
In this section, we first describe our data collection procedure.
Then we present an evaluation of %
the collected data and learned model.\footnote{The code and data for the experiments can be found at \url{https://github.com/YilunZhou/path-naturalness-prediction}}

\subsection{Data Collection}
\label{datacollection}
For our experiments, we used Amazon Mechanical Turk to collect pairwise human rating data. Each questionnaire consisted of 73 pairs, including 60 genuine pairs and 13 quality-control pairs. The quality-control pairs consisted of an obviously  good path and an obviously bad path. The good paths were manually specified and meant to be very natural and straightforward (e.g., Email $\myrightarrow{UsedFor}$ Communication $\myleftarrow{UsedFor}$ Telephone). To generate bad paths, words and relations were sampled independently at random (e.g., ``Beautiful $\myrightarrow{IsA}$ Other $\mydoublearrow{RelatedTo}$ Them $\myleftarrow{MotivatedBy}$ Rug''). %
We only used answers from questionnaires for which all 13 quality-control questions were answered correctly. The genuine pairs consisted of randomly sampled paths from ConceptNet. We used a number of different strategies to sample these paths, as outlined below. Overall, around 1,500 valid responses were collected, giving us around 90,000 pairs of paths to use in different settings of the experiment.

\subsection{Inter-Annotator Agreement}
\label{multiassign_sec}
The human annotators were provided with a generic question: ``which of the two paths is more natural?'' Different participants may interpret this question slightly differently, focusing on different aspects of naturalness. Thus, one important question is how much humans agree with each other about naturalness.
Establishing the inter-annotator agreement level is also useful because it serves as an intrinsic, model-agnostic performance upper bound.

Our default data collection strategy is to obtain only one answer for each pair of paths. This strategy maximizes the coverage of our dataset, while still allowing the model to make probabilistic predictions, as discussed in Section \ref{training}. However, to evaluate the inter-annotator agreement, we also collected a multi-response dataset, with 59 questions each answered by 16 different participants.

Table \ref{multi-assign} summarizes the results of an analysis of this dataset. The first column lists the different possible opinion splits for a given question; for example, 13/3 corresponds to a question for which 13 people preferred one of the paths, and 3 people preferred the other. The second column shows how many of the 59 questions have a given opinion split.

The last two columns of Table \ref{multi-assign} refer to results obtained by our model trained in the 1st setting described below in Section \ref{pred_perf}. The third column shows the number of questions predicted correctly, i.e.\ the number of questions for which the prediction of our model coincides with the majority opinion of the human participants. The last column is the model's average confidence in the majority consensus. The 8/8 row does not include prediction statistics because there is no majority consensus for these questions.

A number of conclusions can be drawn from these results. First, given the disagreement levels shown in the second column, the theoretically best performing model, which always predicts the majority preference, would achieve 70.1\% accuracy on this dataset. This shows that while people do not agree with each other on all of the questions, for the majority of pairs, human annotators do have a clear preference. Note that we should not expect a perfect agreement, since in some cases both of the paths may be equally natural or equally unnatural. This is illustrated in Table \ref{split}, which shows examples of pairs with different opinion splits. As can be seen from these examples, the fact that there is no majority for a given pair does not necessarily mean that the paths involved are of low quality. It simply means that the two paths are approximately equally natural.

As can be seen from the third column in Table \ref{multi-assign}, when there is a clear human consensus about which of the two paths is most natural, our model can predict this with a very high accuracy; e.g.\ for all questions where the opinion split was 14/2 or better, the majority view was predicted correctly. Furthermore, there is a strong positive correlation between the confidence scores predicted by our model, shown in the last column, and the amount of disagreement among human annotators. Our model is thus able to distinguish between cases where the difference in naturalness is clear-cut and cases where human annotators would be undecided.

\begin{table}[!htb]
  \centering
  \begin{tabular}{cccc}
    \toprule
    Opinion Split & \# Q & \# Correct & Avg Conf \\\midrule
    8/8  & 7  & NA & NA     \\
    9/7  & 11 & 5  & 52.2\% \\
    10/6 & 6  & 5  & 60.1\% \\
    11/5 & 8  & 7  & 64.4\% \\
    12/4 & 11 & 7  & 56.6\% \\
    13/3 & 5  & 4  & 70.0\% \\
    14/2 & 5  & 5  & 72.5\% \\
    15/1 & 4  & 4  & 76.2\% \\
    16/0 & 2  & 2  & 89.8\% \\
    \bottomrule
  \end{tabular}
  \normalsize
  \caption{Results for the multi-response dataset. }\label{multi-assign}
\end{table}

\begin{table}[!htb]
    \centering
    \resizebox{\columnwidth}{!}{
    \begin{tabular}{c|c}\toprule
        OS & Paths (The one on top is favored by the majority) \\\midrule
        \multirow{5}{*}{8/8} & Forest $\myleftarrow{AtLocation}$ Country $\mydoublearrow{RelatedTo}$ Geography $\mydoublearrow{RelatedTo}$ Land\\
        & Kingdom $\mydoublearrow{Synonym}$ Land $\myleftarrow{MadeOf}$ Mountain $\mydoublearrow{RelatedTo}$ Hill\\\cmidrule{2-2}
         & Dream $\mydoublearrow{RelatedTo}$ State $\myleftarrow{AtLocation}$ City $\mydoublearrow{RelatedTo}$ Population\\
        & Desk $\mydoublearrow{RelatedTo}$ Office $\mydoublearrow{RelatedTo}$ Area $\mydoublearrow{RelatedTo}$ Science\\\midrule
        \multirow{2}{*}{11/5} & Blood $\myrightarrow{AtLocation}$ Person $\mydoublearrow{RelatedTo}$ Home $\mydoublearrow{Antonym}$ Street\\
        & Range $\mydoublearrow{RelatedTo}$ Food $\myrightarrow{AtLocation}$ Home $\mydoublearrow{Antonym}$ Office\\\midrule
        \multirow{2}{*}{12/4} & Kingdom $\mydoublearrow{RelatedTo}$ Country $\myleftarrow{AtLocation}$ City $\myleftarrow{AtLocation}$ Office\\
        & Sun $\mydoublearrow{RelatedTo}$ Source $\mydoublearrow{RelatedTo}$ Blood $\mydoublearrow{RelatedTo}$ Tissue\\\midrule
        \multirow{2}{*}{13/3} & Person $\mydoublearrow{RelatedTo}$ Woman $\mydoublearrow{DistinctFrom}$ Men $\mydoublearrow{RelatedTo}$ People\\
        & Type $\mydoublearrow{RelatedTo}$ Unit $\mydoublearrow{RelatedTo}$ Card $\mydoublearrow{RelatedTo}$ Holiday\\\midrule
        \multirow{5}{*}{16/0} & School $\myrightarrow{AtLocation}$ City $\myleftarrow{IsA}$ Town $\myrightarrow{AtLocation}$ Country\\
        & Point $\mydoublearrow{RelatedTo}$ Mountain $\mydoublearrow{RelatedTo}$ Wave $\myrightarrow{IsA}$ Woman\\\cmidrule{2-2}
         & People $\mydoublearrow{RelatedTo}$ Life $\myleftarrow{PartOf}$ Fun $\myleftarrow{IsA}$ Soccer\\
        & Plane $\mydoublearrow{Antonym}$ Point $\mydoublearrow{RelatedTo}$ Hand $\mydoublearrow{RelatedTo}$ Instrument\\\bottomrule
    \end{tabular}
    }
    \caption{Pairs of paths with different opinion splits (OS). In each cell the most favored path is shown on top. }
    \label{split}
\end{table}

\subsection{Prediction Accuracy}
\label{pred_perf}
\subsubsection{Experimental Setup}

In this set of experiments we studied how well our model can predict human judgements in three settings.

\begin{itemize}
    \item In the 1st setting, we first determined a set of 100 words. To this end, starting from the center word ``science'', we performed a random walk on ConceptNet, considering only elementary-school-level nouns\footnote{\url{http://www.k12reader.com/}}, until 100 different words had been sampled. We then sampled pairs of paths with 4 or less nodes in the subgraph induced by these 100 words. This dataset included about 10,000 paths. The training set contains 40,000 pairs sampled from 8,000 paths, and the test set contains 1,000 pairs put together from the remaining 2,000 paths. This evaluation allows us to assess the performance of our system in a domain-specific setting, since the model is evaluated on paths involving the same words as those in the training paths (but nonetheless different paths). This performance is recorded in Table \ref{result1}.
    \item In the 2nd setting, we collected another set of 100 words using the same method as above, but using ``money'' as the center word, while additionally ensuring that there is no overlap between this set of 100 words and those from the ``science'' dataset. We then collected 1,000 paths among those 100 words, and put them into 500 pairs, which we use to evaluate models trained on the ``science'' dataset. In this way, we can assess the transfer learning performance of our model. This performance is recorded in Table \ref{result2}.
    \item In the 3rd setting, we selected the nouns, verbs and adjectives from the 5,000 most frequent English words in the Corpus of Contemporary American English\footnote{\url{https://www.wordfrequency.info/}}, for a total of 3,887 words. We then sampled 80,000 paths of up to 5 nodes, which we split into 20,000 training pairs and 20,000 testing pairs (ensuring that there are no overlapping paths between training and testing sets).  Compared with the 1st setting, this represents much wider coverage of the set of words (i.e., embedding space), but much sparser coverage of paths, allowing us to assess the performance of our model in an open-domain setting. This performance is recorded in Table \ref{result3}.
\end{itemize}
\noindent To put our performance into context, we also considered the following four heuristic baselines, all of which have similar mechanics: computing a score for each path (as does our model), and then selecting the path with the higher score.

\noindent \textbf{Source-Target Baseline (ST-B)} scores paths using the cosine similarity between the embeddings of the source and target words (note that the two paths in a pairwise comparison do not typically start and end with the same words);

\noindent \textbf{Pairwise Baseline (Pair-B)} scores paths using the average cosine similarity of all word pairs connected with an edge in the path;

\noindent \textbf{Resource Flow Baseline (Flow-B)} scores paths using the ``path reliability'' method proposed by \citet{lin2015modeling} with the idea that a path is better if there are less branches out of the path;

\noindent \textbf{Length Baseline (Path-B)} scores paths solely by their length, and favors shorter paths to longer ones. %

We performed a grid search of two hyper-parameters: vertex embedding dimension in $\{2, 10, 50, 100, 300\}$ and path code length in $\{1, 2, 5, 10, 20\}$. For the first hyper-parameter, we used principal component analysis (PCA) for dimensionality reduction.
We also tested one-hot encodings of words for the 1st setting. For the other settings, this one-hot encoding cannot be used, because it cannot generalize to unseen words (2nd setting) or scale to a large number of words (3rd setting).
In addition, Length-B performance is only available for the 3rd setting, because for the first two settings we did not explicitly control for path length. As a result, the vast majority of paths are of the cutoff length of 4, and path length is not predictive.

\begin{table}[!htb]
  \centering
  \begin{tabular}{cc|ccccc}\toprule
    & & \multicolumn{5}{c}{Path Code Length}\\
    & & 1 & 2 & 5 & 10 & 20\\\midrule
    \multirow{5}{*}{\rotatebox[origin=c]{90}{Emb.\ Dim.\ }}
    & 2   & 62.1 & 63.1 & 65.8 & 65.3 & 65.6\\%\cmidrule{2-7}
    & 10  & 63.6 & 65.5 & 65.6 & 67.5 & 66.4\\%\cmidrule{2-7}
    & 50  & 63.7 & 65.5 & 66.9 & 67.4 & 67.7\\%\cmidrule{2-7}
    & 100 & 65.4 & 65.9 & 66.1 & \textbf{68.2} & 67.6\\%\cmidrule{2-7}
    & 300 & 66.2 & 66.6 & 67.6 & 67.6 & 67.7\\\midrule
    \multicolumn{2}{c|}{One-Hot} & 67.0 & 66.4 & 67.0 & 67.5 & 67.9\\\midrule
    \multicolumn{2}{c|}{ST-B} &   \multicolumn{5}{c}{53.4}\\\midrule
    \multicolumn{2}{c|}{Pair-B} & \multicolumn{5}{c}{58.0}\\\midrule
    \multicolumn{2}{c|}{Flow-B} & \multicolumn{5}{c}{51.6}\\\midrule
  \end{tabular}
  \normalsize
  \caption{Test accuracy (in percentage) in the domain-specific setting of ``science'' related words.%
  }\label{result1}
\end{table}

\begin{table}[!htb]
  \centering
  \begin{tabular}{cc|ccccc}\toprule
    & & \multicolumn{5}{c}{Path Code Length}\\
    & & 1 & 2 & 5 & 10 & 20\\\midrule
    \multirow{5}{*}{\rotatebox[origin=c]{90}{Emb.\ Dim.\ }}
    & 2   & 60.5 & 64.6 & 63.9 & 65.1 & \textbf{65.5}\\%\cmidrule{2-7}
    & 10  & 60.1 & 61.7 & 64.8 & 64.9 & 64.8\\%\cmidrule{2-7}
    & 50  & 62.5 & 62.8 & 63.1 & 63.3 & 64.0\\%\cmidrule{2-7}
    & 100 & 60.0 & 61.8 & 63.1 & 64.2 & 64.8\\%\cmidrule{2-7}
    & 300 & 58.8 & 60.5 & 61.1 & 62.0 & 62.1\\\midrule
    \multicolumn{2}{c|}{ST-B} &   \multicolumn{5}{c}{53.9}\\\midrule
    \multicolumn{2}{c|}{Pair-B} & \multicolumn{5}{c}{55.9}\\\midrule
    \multicolumn{2}{c|}{Flow-B} & \multicolumn{5}{c}{52.5}\\\midrule
  \end{tabular}
  \normalsize
  \caption{Test accuracy (in percentage) in the transfer learning setting, where the model was trained on the ``science'' dataset and evaluated on the ``money'' dataset. %
  }\label{result2}
\end{table}

\begin{table}[!htb]
  \centering
  \begin{tabular}{cc|ccccc}\toprule
    & & \multicolumn{5}{c}{Path Code Length}\\
    & & 1 & 2 & 5 & 10 & 20\\\midrule
    \multirow{5}{*}{\rotatebox[origin=c]{90}{Emb.\ Dim.\ }}
    & 2   & 61.4 & 60.5 & 63.1 & 62.9 & 63.3\\%\cmidrule{2-7}
    & 10  & 61.2 & 62.6 & 65.0 & 64.9 & 64.2\\%\cmidrule{2-7}
    & 50  & 61.6 & 64.5 & 64.8 & 65.4 & 65.8\\%\cmidrule{2-7}
    & 100 & 62.0 & 64.3 & 65.0 & \textbf{67.9} & 65.1\\%\cmidrule{2-7}
    & 300 & 61.9 & 63.0 & 65.7 & 65.4 & 65.4\\\midrule
    \multicolumn{2}{c|}{ST-B} &   \multicolumn{5}{c}{52.3}\\\midrule
    \multicolumn{2}{c|}{Pair-B} & \multicolumn{5}{c}{57.5}\\\midrule
    \multicolumn{2}{c|}{Flow-B} & \multicolumn{5}{c}{46.8}\\\midrule
    \multicolumn{2}{c|}{Length-B} & \multicolumn{5}{c}{55.8}\\\bottomrule
  \end{tabular}
  \normalsize
  \caption{Test accuracy (in percentage) in the open-domain setting of the most frequent English words.%
  }\label{result3}
\end{table}

\subsubsection{Quantitative Analysis}
For the first setting (Table \ref{result1}), we can see that performance was relatively insensitive to embedding dimension and code length, as long as both were sufficiently high. However, performance on the 2nd setting (Table \ref{result2}) dropped significantly with higher embedding dimensions, suggesting an overfitting problem, with an embedding dimension of 2 achieving best generalization performance. This suggests that a model that was trained on one domain can indeed successfully be applied to another domain, as long as the embedding dimension is small enough to allow for sufficient generalization.
In the 3rd setting (Table \ref{result3}), we found that a higher embedding dimension is necessary, most likely because the range of meanings is more diverse for this dataset, which includes not only nouns, but also verbs and adjectives.

For all three settings, considering the level of agreement found in Section \ref{multiassign_sec}, we found that the performance of our model approaches the expected performance upper bound. Moreover, our model performs substantially and consistently better than all of the baselines. In fact, in Table \ref{result3} we can see that only one of the baselines is able to outperform the simple path length heuristic (Length-B), and only in a minimal way.

\subsubsection{Qualitative Analysis}
\label{qualitative}
To qualitatively compare how our model prediction differs from the baselines, Table \ref{3compare} presents the best paths between the words ``health'' and ``computer'' (of up to 4 nodes) selected by our method and the two strongest baselines.

\begin{table}[!htb]
  \centering
  \resizebox{\columnwidth}{!}{
  \begin{tabular}{c|l}\toprule
    \multirow{5}{*}{\rotatebox[origin=c]{90}{Pairwise}}
    & Health $\mydoublearrow{RelatedTo}$ Care $\myrightarrow{IsA}$ Work $\mydoublearrow{RelatedTo}$ Computer \\\cmidrule{2-2}
    & Health $\mydoublearrow{RelatedTo}$ Care $\mydoublearrow{RelatedTo}$ Help $\mydoublearrow{RelatedTo}$ Computer \\\cmidrule{2-2}
    & Health $\mydoublearrow{RelatedTo}$ Care $\mydoublearrow{RelatedTo}$ Do $\myleftarrow{CapableOf}$ Computer \\\cmidrule{2-2}
    & Health $\mydoublearrow{RelatedTo}$ Care $\myleftarrow{Desires}$ Person $\myrightarrow{Desires}$ Computer \\\bottomrule\toprule
    \multirow{5}{*}{\rotatebox[origin=c]{90}{Length}}
    & Health $\mydoublearrow{RelatedTo}$ System $\mydoublearrow{RelatedTo}$ Computer \\\cmidrule{2-2}
    & Health $\mydoublearrow{RelatedTo}$ Level $\mydoublearrow{RelatedTo}$ Computer \\\cmidrule{2-2}
    & Health $\mydoublearrow{RelatedTo}$ Apple $\mydoublearrow{RelatedTo}$ Computer \\\cmidrule{2-2}
    & Health $\mydoublearrow{RelatedTo}$ Well $\mydoublearrow{RelatedTo}$ Screw $\myrightarrow{AtLocation}$ Computer \\\bottomrule\toprule
    \multirow{5}{*}{\rotatebox[origin=c]{90}{Our Model}}
    & Health $\myleftarrow{Antonym}$ Disease $\myleftarrow{Causes}$ Virus $\myrightarrow{AtLocation}$ Computer \\\cmidrule{2-2}
    & Health $\mydoublearrow{RelatedTo}$ Care $\myrightarrow{IsA}$ Work $\mydoublearrow{RelatedTo}$ Computer \\\cmidrule{2-2}
    & Health $\mydoublearrow{RelatedTo}$ Insurance $\myrightarrow{HasProperty}$ Expensive $\myleftarrow{HasProperty}$ Computer \\\cmidrule{2-2}
    & Health $\rightarrow$ Sickness $\rightarrow$ Virus $\myrightarrow{AtLocation}$ Computer \\\bottomrule
  \end{tabular}
  }
  \caption{Best paths of 4 or less nodes between ``health'' and ``computer'' ranked by baselines and our model.} \label{3compare}
\end{table}

We can see that the top paths found by the pairwise baseline lack variety, most likely due to the high embedding similarity between ``health'' and ``care''. In addition, the length baseline does not perform satisfactorily because the shortest paths are not easily understandable without more explanation, with nearly all relations being the generic ``$\mydoublearrow{RelatedTo}$''. By contrast, our model (trained on the 3rd open-domain setting) selects paths that overcome both drawbacks. It covers a wider variety of concepts such as ``virus'' and ``expensive'' and favors longer paths with smoother transition of node semantics.

In another evaluation, we inspect paths for which our model and the pairwise baseline significantly disagree. We randomly sampled 120,000 paths from the entire ConceptNet graph and use the two methods to predict their quality. Table \ref{nat_top_pair_bot} lists paths that are predicted to be among the top 10\% most natural by our model but are in the bottom 10\% according to the pairwise baseline. Table \ref{nat_bot_pair_top} conversely shows paths that are among the 10\% least natural according to our model but the 10\% most natural according to the baselines. Paths in both tables were selected at random.

\begin{table}%
  \centering
  \resizebox{\columnwidth}{!}{
  \begin{tabular}{l}\toprule
    Lead $\myrightarrow{HasProperty}$ Toxic $\mydoublearrow{RelatedTo}$ Toxicant $\mydoublearrow{Synonym}$ Poison \\\midrule
    Wave $\myrightarrow{IsA}$ Fluctuation $\mydoublearrow{RelatedTo}$ Brainwave $\myrightarrow{DerivedFrom}$ Brain \\\midrule
    Space $\myleftarrow{MadeOf}$ Passageway $\myrightarrow{AtLocation}$ Building $\mydoublearrow{RelatedTo}$ Station \\\midrule
    Food $\mydoublearrow{RelatedTo}$ Hechsher $\mydoublearrow{RelatedTo}$ Kashrut $\mydoublearrow{RelatedTo}$ Law \\\bottomrule
  \end{tabular}
  }
  \caption{Paths predicted to be among 10\% most natural by our model, but 10\% least natural by the pairwise baseline. } \label{nat_top_pair_bot}
\end{table}

\begin{table}%
  \centering
  \begin{tabular}{l}\toprule
    Bee $\mydoublearrow{RelatedTo}$ A $\mydoublearrow{RelatedTo}$ After $\mydoublearrow{RelatedTo}$ Attack \\\midrule
    Space $\mydoublearrow{RelatedTo}$ Very $\mydoublearrow{RelatedTo}$ Little $\mydoublearrow{RelatedTo}$ Moon \\\midrule
    Lead $\mydoublearrow{Synonym}$ Run $\mydoublearrow{RelatedTo}$ Very $\mydoublearrow{RelatedTo}$ Poison \\\midrule
    Adult $\mydoublearrow{RelatedTo}$ A $\mydoublearrow{RelatedTo}$ The $\mydoublearrow{RelatedTo}$ English \\\bottomrule
  \end{tabular}
  \caption{Paths predicted to be among 10\% most natural by the pairwise baseline, but 10\% least natural by our model. } \label{nat_bot_pair_top}
\end{table}

As we can see, the paths in Table \ref{nat_top_pair_bot} are intuitively quite meaningful, with all paths using some rather uncommon but relevant words, which is perhaps clearest in the last example (Hechsher refers to a certification given to a food product that complies with Jewish religious law and Kashrut refers to the set of Jewish dietary laws). By comparison, the paths in Table \ref{nat_bot_pair_top} mostly use very common but uninformative words. This analysis suggests that the pairwise baseline suffers from the so-called hubness problem, i.e.\ the problem that in high-dimensional vector space embeddings, there are typically a few central objects which are highly similar to many of the other objects \cite{radovanovic2010hubs}, an issue which is known to affect word embeddings \cite{dinu2014improving}. The pairwise baseline favors paths with very common words, even if they are not semantically related, because these words act as hubs in the word embedding.

\subsection{Ablation Study on Feature Importance}
To discern the contribution of each feature to the prediction performance, we trained variants of our model with subsets of the full feature set. For this analysis, we used the 3rd evaluation setting with the best performing hyper-parameters (embedding dimension of 100 and path code length of 10). The results are summarized in Table \ref{ablation}.

\begin{table}[htb]
  \centering
  \begin{tabular}{lc}\toprule
  All Features & 67.9\%\\\midrule
    No vertex embedding & 65.4\% \\
    No vertex frequency & 67.3\% \\
    No vertex degree & 67.1\% \\
    No edge ends similarity & 63.5\%\\
    No edge direction & 67.8\%\\
    No edge relation & 67.3\%\\
    No edge provenance & 67.2\%\\
    No vertex and edge sense scores & 66.9\%\\\midrule
    Vertex features only & 61.6\%\\
    Edge features only & 63.7\%\\\bottomrule
  \end{tabular}
  \normalsize
  \caption{Ablation study on feature importance. }\label{ablation}
\end{table}

We can see that leaving out the edge ends similarity feature hurts performance most. %
This is also in accordance with the fact that pairwise baseline in Table \ref{result3}, which only uses this feature, performs best among the four baseline methods. Furthermore, the second-highest reduction is found when leaving out the vertex embedding feature. Finally, we see that neither vertex features nor edge features alone can achieve optimal prediction performance.

\subsection{Naturalness as a Path Selection Criterion}
Next we discuss three experiments which are aimed at assessing how well naturalness indicates path quality. %
We start with an intrinsic evaluation of the semantic coherence of natural paths, after which we discuss two extrinsic tasks: information retrieval and analogy inference.

\subsubsection{Semantic Coherence}
Let us define the \emph{type} of a path as the sequence of relations appearing in that path, e.g.\ the type of the path ``A $\myrightarrow{IsA}$ B $\myrightarrow{HasA}$ C'' is ($\myrightarrow{IsA}$, $\myrightarrow{HasA}$).
In this experiment, we consider pairs of nodes in ConceptNet that are connected with a relational path of a given type, as well as a one-step relation (e.g., ``A $\myrightarrow{HasA}$ C''). In this situation, we call the latter relation a \em path-summarizing (PS) relation \rm for the considered path type. %
Many path types uniquely determine the PS relation, or at least narrow down the set of candidate PS relations to a small number. For instance, for a path of type ($\myrightarrow{IsA}$, $\myrightarrow{HasA}$) we would expect to only see $\myrightarrow{HasA}$ as the PS relation. However, due to the presence of noise in ConceptNet, some relations of this type will actually have a different PS relation. Motivated by this view, as an intrinsic evaluation task, we propose to assess the extent to which a path selection method is able to select \emph{semantically coherent} paths in terms of the variability of the PS relations among the selected paths.

In particular,
for a set of paths, we first determine a mapping from path types to counters of PS relations: $\{p_1\allowbreak\rightarrow \allowbreak[(r_{11}, c_{11}),\allowbreak(r_{12}, c_{12}),\allowbreak...,\allowbreak(r_{1n}, c_{1n})],\allowbreak...,\allowbreak p_m\rightarrow \allowbreak[(r_{m1}, c_{m1}),\allowbreak(r_{m2}, c_{m2}),\allowbreak...,\allowbreak(r_{mn}, \allowbreak c_{mn})]\}$, in which each $p_i$ is a path type, each $r_{ij}$ is one of 46 relation types from ConceptNet, and each $c_{ij}$ is the number of times the relation $r_{ij}$ was found as a PS relation for path type $p_i$.
The average entropy of the PS relations is then defined as:
\begin{align*}
\frac{-\sum_{i=1}^{m}\left[C_i\cdot\sum_{j=1}^{n}p_{ij}\ln(p_{ij})\right]}{C},
\end{align*}
where $C_i=\sum_{j=1}^nc_{ij},\allowbreak C=\sum_{i=1}^{m}C_i,\allowbreak p_{ij}=c_{ij}/C_i$. A higher average entropy suggest a higher proportion of spurious PS relations and thus less semantically coherent paths.

\begin{figure}
  \centering
  \includegraphics[width=2.3in]{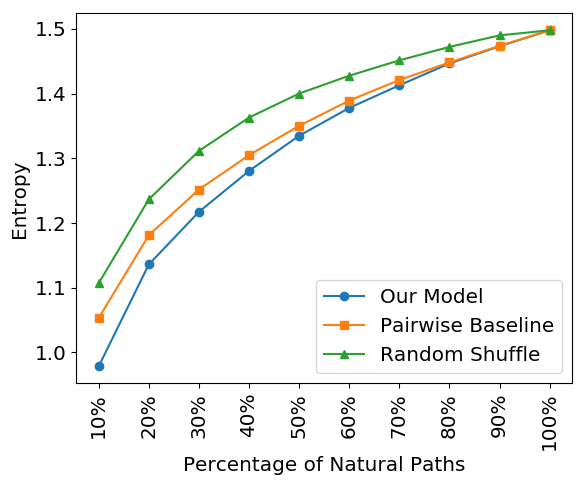}
  \caption{Average entropy of path-summarizing relations for different proportions of natural paths.}\label{entropy}
\end{figure}

Using the 3,887 words from the 3rd setting, we generated 126,600 length-4 paths whose source and target nodes are also directly connected by a relation (i.e. the PS relation). Our model was applied to rank the paths by naturalness. Based on this ranking, we calculate the average entropy of top $N\%$ natural paths, where $N$ varies from 10 to 100, and plot the result in Figure \ref{entropy}. %
We compared our model to Pair-B, the best-performing baseline for predicting naturalness. The entropy for a random shuffle of the dataset is also depicted for comparison. We see that using our model to select the most natural paths consistently leads to the lowest entropy. Note that all three methods will converge to the same entropy at the 100\% mark, at which point all paths are used, regardless of the path selection method. %

\subsubsection{Information Retrieval}
When using a search engine, users often provide under-specified queries,
e.g.\ because they assume that the search engine is ``smart'' enough to infer their true intention or because they do not actually have a clear idea on how to formulate a better query. Moreover, even if a query is unambiguous, it may use different terms than those appearing in the most relevant documents. To deal with such issues, many information retrieval systems rely on some form of query expansion \citep{Xu:1996:QEU:243199.243202}, a strategy for automatically adding semantically related keywords to the user's query.
In this experiment, we study the performance of a query expansion method based on natural ConceptNet paths. %

We use the TREC 2004 Robust Retrieval Track dataset \citep{voorhees2005overview}, commonly known as ROBUST04, consisting of 250 queries and around 260,000 documents. Each query from this dataset typically consists of 2 or 3 words, and is associated on average with 1,000 annotated documents (marked as either positive or negative). We use a simple TF-IDF retrieval model, in which the TF-IDF weighting scheme is used to vectorize both the documents and the queries and the cosine similarity is used as the ranking metric. Following the work by \citet{kotov2012tapping}, we focus on hard queries, which are defined as those with Precision at 10 (P@10) performance of 0 when only using the original query terms. We then evaluate how the performance can be improved by expanding the query with terms that are selected using a given strategy.
There are 44 examples of such hard queries among the original 250 queries.

To expand queries based on ConceptNet, we first find all paths, consisting of up to 4 nodes, which connect any two of the query terms. For example, if the query is ``drug crime organization'', we find paths between the pairs (drug, crime), (drug, organization), and (crime, organization).
We then add the words from the most natural paths to the original query. In particular, we add a fixed number of words by processing the paths in the order defined by the chosen path ranking method. %
To this end, we have considered the following strategies for ranking paths:
\begin{enumerate}
    \item naturalness score, which is calculated from our model trained on the open-domain (3rd) setting;
    \item pairwise score, which is the pairwise baseline;
    \item length score, which is the length baseline (i.e. shorter paths are better), with random tie-breaking;
    \item random score, which assigns a random score to each path;
    \item naturalness+length score, which first favors shorter paths, but uses our naturalness score to rank paths with same length.
\end{enumerate}
To put our results into context, we also compared this method with a query expansion strategy based on the GloVe word embedding. Specifically, in this case we add the words that are closest to the query words, ordered by their cosine similarity in the vector space. Note that each of the considered strategies will yield an ordered list of possible words to add to the query. We study the retrieval performance of using different cut-offs in this list.

\begin{figure}
    \centering
    \includegraphics[width=2.5in]{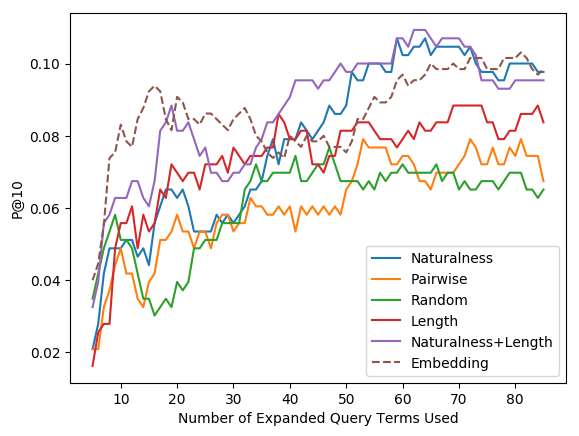}
    \includegraphics[width=2.5in]{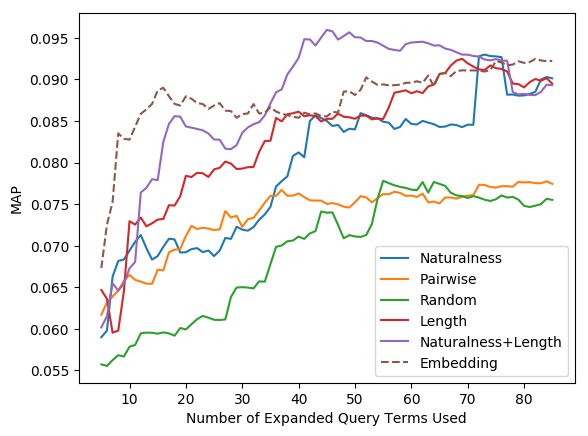}
    \caption{Performance of query expansion for hard queries in terms of Precision at 10 (P@10, top) and Mean Average Precision (MAP, bottom).}
    \label{ir_figure}
\end{figure}

The P@10 and MAP performances are shown in Figure \ref{ir_figure}. We can see that for both evaluation metrics, the naturalness+length score achieves the highest performance, suggesting that while nodes in shorter paths are generally more relevant to original queries, considerable gains can be made by using naturalness to rank paths of the same length. The word embedding model also performs well, but it cannot reach the best overall performance of the two naturalness based methods.

\subsubsection{Analogy Inference}
\citet{boteanu15solving} used relation paths to solve analogy questions of the following form:
 \begin{verbatim}
   Question: Dog:Animal::
          A: Table:Chair
          B: Apple:Fruit
          C: Fast:Slow
          D: Ice:Cold
 \end{verbatim}
The goal is to select the word pair which has the same relation as the query pair (e.g.\ the correct answer in the question above is $B$ because dog is a type of animal, and apple is a type of fruit). %
In their proposed method, the quality of the analogy $a:b::c:d$ (meaning $a$ is to $b$ as $c$ is to $d$) is calculated using the proportion of overlapping relations on paths connecting $(a, b)$ and $(c, d)$ respectively. \citet{boteanu15solving} considered paths of two (i.e.\ direct connection) and three (i.e.\ with an intermediate hop) nodes.

\begin{figure}[!htb]
  \centering
  \includegraphics[width=2.5in]{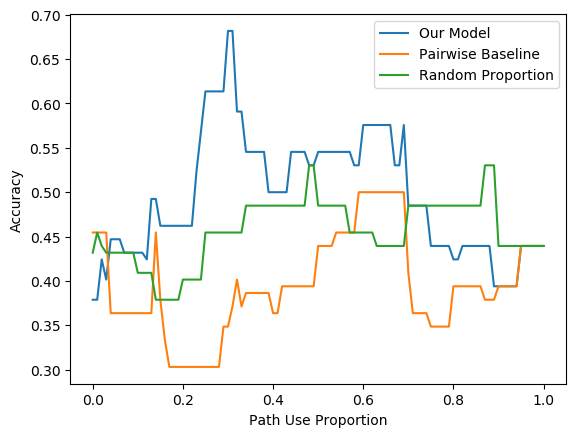}
  \caption{Performance on the 11 analogy questions for which direct connections are not available, using various proportions of natural paths. }\label{progressive}
\end{figure}

We found that when direct connections exist between words in the query pair, using only these direct connection (and not including 3-node paths) achieves the best performance. However, not all query pairs have direct connections, in which case we are required to use 3-node paths. For our dataset of 85 questions, 74 have associated direct edge connections. Thus, we study the effect of only considering the most natural paths, rather than all of them, on the remaining 11 questions.
Figure \ref{progressive} shows the performance of the prediction as a function of the proportion of 3-node paths used. Using our model to filter 3-node paths, the best performance is obtained when only the 30\% most natural paths are considered. As a comparison, the pairwise baseline is not able to outperform a random selection.

Across all three evaluations, we consistently find that including all (or too many) paths leads to sub-optimal performance, due to the inclusion of noisy paths, and that our proposed method outperforms existing heuristics based on word embedding and path length.

\section{Conclusions}

Our work is motivated by the observation that commonsense knowledge graphs such as ConceptNet are noisy and informal, and that many of the relational paths in such knowledge graphs are therefore non-sensical. This means that the only way in which applications can successfully make use of relational paths from such resources is by relying on a method to filter out these non-sensical paths. While this problem has already been studied by others, previous solutions relied on relatively simple heuristics to select higher-quality paths, which we found to perform poorly in practice. In fact, in the extrinsic evaluation tasks, we found that some heuristics were not even able to outperform random selection (Figures \ref{ir_figure} and \ref{progressive}).  %

Rather than engineering more sophisticated heuristics, the solution we proposed in this paper is to take a data-driven approach and learn the concept of path \em naturalness \rm based on crowdsourced judgments. The main insights we obtained are as follows:
\begin{enumerate}
    \item although we intentionally left the interpretation of naturalness open to crowdsourcers, they do largely share a common perception of this concept;
    \item %
    this concept can be effectively learned by an LSTM model, with accuracy close to an empirical upper bound derived from the inter-annotator agreement level; and
    \item our learned model outperforms previously proposed heuristics as a path selection method in several intrinsic and extrinsic evaluation tasks.
\end{enumerate}

\section*{Acknowledgment}
Steven Schockaert was supported by ERC Starting Grant 637277.

\ifextended

\section*{Appendix: Sense Score Computation}
Since ConceptNet does not distinguish between senses of a given word, we calculate this feature in order to capture our intuition that paths mixing different senses of the same word will be perceived as less natural; for example, ``Knowledge $\myleftarrow{HasA}$ Book $\mydoublearrow{RelatedTo}$ Restaurant'' may be nonintuitive since the word ``book'' is used in two different senses: ``something to read'' in the first part, and ``reserve (accommodations, a place, etc.)'' in the second. Moreover, not all senses are ``equidistant'' from each other; for example, the second part of the path ``Knowledge $\myleftarrow{HasA}$ Book $\myleftarrow{IsA}$ Notebook'' incorporates yet another sense of ``book'': ``a collection of blank pages to write on.'' However, the change in meaning is much less significant in this path, and as a result this path is perceived as more natural.

\citet{chen2011combining} modeled a sense by its word sense profile (WSP). Using WordNet, the WSP of a \em sense \rm (e.g., first sense of the word ``car'') is the union of the following sets of words (words in parentheses are for the ``car'' example):
\begin{enumerate}
  \itemsep0em
  \item words in the synonym set (auto, machine, motorcar)
  \item IsA relation (ambulance, bus, cab, etc.)
  \item HasA relation (accelerator, bumper, roof, etc.)
  \item definition excluding stop words (\st{a} motor vehicle \st{with} four wheels usually propelled \st{by} \st{an} internal combustion engine)
\end{enumerate}

\citet{chen2011combining} defined the similarity between \em two words \rm $f_{ww}\allowbreak(w_1, \allowbreak w_2)$ as the Normalized Google Distance \citep{cilibrasi2007google}. However, because Google now rate-limits their query interface, we instead use cosine similarity of embeddings as the similarity function.

The similarity function of \em a word and a sense \rm is defined as:
\begin{align*}
f_{ws}(w, s)=\frac{\sum_{w'\in WSP(s)}f_{ww}(w, w')}{|WSP(s)|}.
\end{align*}
In practice, when taking the average, we only use 10 $w'\in $ \textit{WSP(s)} that are most similar to $w$ according to $f_{ww}$.

For a path consisting of words $w_1, w_2, ..., w_n$, we try to determine the most plausible sense for each word. Specifically, for each word $w_i$, we find the sense $s^{w_i}$ for $w_i$ that maximizes $f_{ws}(w_{i-1}, s^{w_i})\allowbreak+\allowbreak f_{ws}\allowbreak(\allowbreak w_{i+1},\allowbreak s^{w_i})\allowbreak]\allowbreak /\allowbreak 2$. For boundary words, the only neighbor is used.

\paragraph{Vertex sense score}

Given a disambiguated path, we then calculate the sense score for each vertex and edge. The assigned sense for $w_i$ is denoted as $s_i^*$, and the sense score for the vertex $v_i$ is:
\begin{align*}
\frac{f_{ws}(w_{i-1}, s_i^*)+f_{ws}(w_{i+1}, s_i^*)}{\max_{s_i'}f_{ws}(w_{i-1}, s_i')+\max_{s_i'}f_{ws}(w_{i+1}, s_i')}.
\end{align*}
Note that the maximum sense score of 1 is attained when the sense $s_i^*$ is the sense most closely related to both $w_{i-1}$ and $w_{i+1}$. This is intuitively the case if the edges $(w_{i-1},w_i)$ and $(w_i,w_{i+1})$ are based on the same sense of word $w_i$. Also note that the sense score for the first and last vertices is 1 by design.

\paragraph{Edge sense score}

Using the same notation as vertex sense, the sense score for the edge $e_i=(v_{i-1}, v_i)$ is:
\begin{align*}
\frac{f_{ws}(w_{i-1}, s_i^*)+f_{ws}(w_{i}, s_{i-1}^*)}{\max_{s_i'}f_{ws}(w_{i-1}, s_i')+\max_{s_{i-1}'}f_{ws}(w_{i}, s_{i-1}')}.
\end{align*}

\fi

\bibliographystyle{ACM-Reference-Format}
\balance
\bibliography{references}


\begin{thebibliography}{36}

%%% the Web address.

\ifx \showCODEN    \undefined \def \showCODEN     #1{\unskip}     \fi
\ifx \showDOI      \undefined \def \showDOI       #1{#1}\fi
\ifx \showISBNx    \undefined \def \showISBNx     #1{\unskip}     \fi
\ifx \showISBNxiii \undefined \def \showISBNxiii  #1{\unskip}     \fi
\ifx \showISSN     \undefined \def \showISSN      #1{\unskip}     \fi
\ifx \showLCCN     \undefined \def \showLCCN      #1{\unskip}     \fi
\ifx \shownote     \undefined \def \shownote      #1{#1}          \fi
\ifx \showarticletitle \undefined \def \showarticletitle #1{#1}   \fi
\ifx \showURL      \undefined \def \showURL       {\relax}        \fi
% invisible to TeX
\providecommand\bibfield[2]{#2}
\providecommand\bibinfo[2]{#2}
\providecommand\natexlab[1]{#1}
\providecommand\showeprint[2][]{arXiv:#2}

\bibitem[\protect\citeauthoryear{Aroyo and Welty}{Aroyo and Welty}{2014}]%
        {aroyo2014three}
\bibfield{author}{\bibinfo{person}{Lora Aroyo} {and} \bibinfo{person}{Chris
  Welty}.} \bibinfo{year}{2014}\natexlab{}.
\newblock \showarticletitle{The Three Sides of {CrowdTruth}}.
\newblock \bibinfo{journal}{\emph{Human Computation}}  \bibinfo{volume}{1}
  (\bibinfo{year}{2014}), \bibinfo{pages}{31--34}.
\newblock


\bibitem[\protect\citeauthoryear{Auer, Bizer, Kobilarov, Lehmann, Cyganiak, and
  Ives}{Auer et~al\mbox{.}}{2007}]%
        {auer2007dbpedia}
\bibfield{author}{\bibinfo{person}{S{\"o}ren Auer}, \bibinfo{person}{Christian
  Bizer}, \bibinfo{person}{Georgi Kobilarov}, \bibinfo{person}{Jens Lehmann},
  \bibinfo{person}{Richard Cyganiak}, {and} \bibinfo{person}{Zachary Ives}.}
  \bibinfo{year}{2007}\natexlab{}.
\newblock \showarticletitle{{DBpedia}: A Nucleus for a Web of Open Data}.
\newblock \bibinfo{journal}{\emph{The Semantic Web}} (\bibinfo{year}{2007}),
  \bibinfo{pages}{722--735}.
\newblock


\bibitem[\protect\citeauthoryear{Bollacker, Evans, Paritosh, Sturge, and
  Taylor}{Bollacker et~al\mbox{.}}{2008}]%
        {bollacker2008freebase}
\bibfield{author}{\bibinfo{person}{Kurt Bollacker}, \bibinfo{person}{Colin
  Evans}, \bibinfo{person}{Praveen Paritosh}, \bibinfo{person}{Tim Sturge},
  {and} \bibinfo{person}{Jamie Taylor}.} \bibinfo{year}{2008}\natexlab{}.
\newblock \showarticletitle{Freebase: A Collaboratively Created Graph Database
  for Structuring Human Knowledge}. In \bibinfo{booktitle}{\emph{Proceedings of
  the 2008 ACM SIGMOD International Conference on Management of Data}}. ACM,
  \bibinfo{pages}{1247--1250}.
\newblock


\bibitem[\protect\citeauthoryear{Boteanu and Chernova}{Boteanu and
  Chernova}{2015}]%
        {boteanu15solving}
\bibfield{author}{\bibinfo{person}{Adrian Boteanu} {and} \bibinfo{person}{Sonia
  Chernova}.} \bibinfo{year}{2015}\natexlab{}.
\newblock \showarticletitle{Solving and Explaining Analogy Questions Using
  Semantic Networks}. In \bibinfo{booktitle}{\emph{AAAI Conference on
  Artificial Intelligence}}. AAAI, \bibinfo{pages}{1460--1466}.
\newblock


\bibitem[\protect\citeauthoryear{Bouraoui, Jameel, and Schockaert}{Bouraoui
  et~al\mbox{.}}{2018}]%
        {DBLP:conf/coling/BouraouiJS18}
\bibfield{author}{\bibinfo{person}{Zied Bouraoui}, \bibinfo{person}{Shoaib
  Jameel}, {and} \bibinfo{person}{Steven Schockaert}.}
  \bibinfo{year}{2018}\natexlab{}.
\newblock \showarticletitle{Relation Induction in Word Embeddings Revisited}.
  In \bibinfo{booktitle}{\emph{Proceedings of the 27th International Conference
  on Computational Linguistics (COLING)}}. \bibinfo{pages}{1627--1637}.
\newblock


\bibitem[\protect\citeauthoryear{Carlson, Betteridge, Kisiel, Settles,
  Hruschka~Jr, and Mitchell}{Carlson et~al\mbox{.}}{2010}]%
        {carlson2010toward}
\bibfield{author}{\bibinfo{person}{Andrew Carlson}, \bibinfo{person}{Justin
  Betteridge}, \bibinfo{person}{Bryan Kisiel}, \bibinfo{person}{Burr Settles},
  \bibinfo{person}{Estevam~R Hruschka~Jr}, {and} \bibinfo{person}{Tom~M
  Mitchell}.} \bibinfo{year}{2010}\natexlab{}.
\newblock \showarticletitle{Toward an Architecture for Never-Ending Language
  Learning}. In \bibinfo{booktitle}{\emph{AAAI Conference on Artificial
  Intelligence}}. \bibinfo{pages}{1306--1313}.
\newblock


\bibitem[\protect\citeauthoryear{Chen and Liu}{Chen and Liu}{2011}]%
        {chen2011combining}
\bibfield{author}{\bibinfo{person}{Junpeng Chen} {and} \bibinfo{person}{Juan
  Liu}.} \bibinfo{year}{2011}\natexlab{}.
\newblock \showarticletitle{Combining {ConceptNet} and {WordNet} for Word Sense
  Disambiguation}. In \bibinfo{booktitle}{\emph{International Joint Conference
  on Natural Language Processing (IJCNLP)}}. \bibinfo{pages}{686--694}.
\newblock


\bibitem[\protect\citeauthoryear{Cilibrasi and Vitanyi}{Cilibrasi and
  Vitanyi}{2007}]%
        {cilibrasi2007google}
\bibfield{author}{\bibinfo{person}{Rudi~L Cilibrasi} {and}
  \bibinfo{person}{Paul~MB Vitanyi}.} \bibinfo{year}{2007}\natexlab{}.
\newblock \showarticletitle{The Google Similarity Distance}.
\newblock \bibinfo{journal}{\emph{IEEE Transaction on Knowledge and Data
  Engineering (TKDE)}} \bibinfo{volume}{19}, \bibinfo{number}{3}
  (\bibinfo{year}{2007}).
\newblock


\bibitem[\protect\citeauthoryear{Das, Neelakantan, Belanger, and McCallum}{Das
  et~al\mbox{.}}{2017}]%
        {das2017chains}
\bibfield{author}{\bibinfo{person}{Rajarshi Das}, \bibinfo{person}{Arvind
  Neelakantan}, \bibinfo{person}{David Belanger}, {and} \bibinfo{person}{Andrew
  McCallum}.} \bibinfo{year}{2017}\natexlab{}.
\newblock \showarticletitle{Chains of Reasoning over Entities, Relations, and
  Text Using Recurrent Neural Networks}. In
  \bibinfo{booktitle}{\emph{Proceedings of the 15th Conference of the European
  Chapter of the Association for Computational Linguistics (EACL)}},
  Vol.~\bibinfo{volume}{1}. \bibinfo{pages}{132--141}.
\newblock


\bibitem[\protect\citeauthoryear{Dinu, Lazaridou, and Baroni}{Dinu
  et~al\mbox{.}}{2015}]%
        {dinu2014improving}
\bibfield{author}{\bibinfo{person}{Georgiana Dinu}, \bibinfo{person}{Angeliki
  Lazaridou}, {and} \bibinfo{person}{Marco Baroni}.}
  \bibinfo{year}{2015}\natexlab{}.
\newblock \showarticletitle{Improving Zero-Shot Learning by Mitigating the
  Hubness Problem}. In \bibinfo{booktitle}{\emph{Proceedings of the
  International Conference on Learning Representations (ICLR)}}.
\newblock


\bibitem[\protect\citeauthoryear{Faruqui, Dodge, Jauhar, Dyer, Hovy, and
  Smith}{Faruqui et~al\mbox{.}}{2015}]%
        {faruqui2014retrofitting}
\bibfield{author}{\bibinfo{person}{Manaal Faruqui}, \bibinfo{person}{Jesse
  Dodge}, \bibinfo{person}{Sujay~Kumar Jauhar}, \bibinfo{person}{Chris Dyer},
  \bibinfo{person}{Eduard Hovy}, {and} \bibinfo{person}{Noah~A. Smith}.}
  \bibinfo{year}{2015}\natexlab{}.
\newblock \showarticletitle{Retrofitting Word Vectors to Semantic Lexicons}. In
  \bibinfo{booktitle}{\emph{Proceedings of the Conference of the North American
  Chapter of the Association for Computational Linguistics: Human Language
  Technologies (NAACL:HLT)}}. Association for Computational Linguistics,
  \bibinfo{pages}{1606--1615}.
\newblock


\bibitem[\protect\citeauthoryear{Gardner, Talukdar, Krishnamurthy, and
  Mitchell}{Gardner et~al\mbox{.}}{2014}]%
        {gardner2014incorporating}
\bibfield{author}{\bibinfo{person}{Matt Gardner}, \bibinfo{person}{Partha
  Talukdar}, \bibinfo{person}{Jayant Krishnamurthy}, {and} \bibinfo{person}{Tom
  Mitchell}.} \bibinfo{year}{2014}\natexlab{}.
\newblock \showarticletitle{Incorporating Vector Space Similarity in Random
  Walk Inference over Knowledge Bases}. In
  \bibinfo{booktitle}{\emph{Proceedings of the Conference on Empirical Methods
  in Natural Language Processing (EMNLP)}}. \bibinfo{pages}{397--406}.
\newblock


\bibitem[\protect\citeauthoryear{Guu, Miller, and Liang}{Guu
  et~al\mbox{.}}{2015}]%
        {guu2015traversing}
\bibfield{author}{\bibinfo{person}{Kelvin Guu}, \bibinfo{person}{John Miller},
  {and} \bibinfo{person}{Percy Liang}.} \bibinfo{year}{2015}\natexlab{}.
\newblock \showarticletitle{Traversing Knowledge Graphs in Vector Space}. In
  \bibinfo{booktitle}{\emph{Proceedings of the Conference on Empirical Methods
  in Natural Language Processing (EMNLP)}}.
\newblock


\bibitem[\protect\citeauthoryear{Hochreiter and Schmidhuber}{Hochreiter and
  Schmidhuber}{1997}]%
        {hochreiter1997long}
\bibfield{author}{\bibinfo{person}{Sepp Hochreiter} {and}
  \bibinfo{person}{J{\"u}rgen Schmidhuber}.} \bibinfo{year}{1997}\natexlab{}.
\newblock \showarticletitle{Long Short-Term Memory}.
\newblock \bibinfo{journal}{\emph{Neural Computation}} \bibinfo{volume}{9},
  \bibinfo{number}{8} (\bibinfo{year}{1997}), \bibinfo{pages}{1735--1780}.
\newblock


\bibitem[\protect\citeauthoryear{Kingma and Ba}{Kingma and Ba}{2015}]%
        {kingma2014adam}
\bibfield{author}{\bibinfo{person}{Diederik Kingma} {and}
  \bibinfo{person}{Jimmy Ba}.} \bibinfo{year}{2015}\natexlab{}.
\newblock \showarticletitle{Adam: A Method for Stochastic Optimization}. In
  \bibinfo{booktitle}{\emph{Proceedings of the International Conference
  Learning Representations (ICLR)}}.
\newblock


\bibitem[\protect\citeauthoryear{Kotov and Zhai}{Kotov and Zhai}{2012}]%
        {kotov2012tapping}
\bibfield{author}{\bibinfo{person}{Alexander Kotov} {and}
  \bibinfo{person}{Chengiang Zhai}.} \bibinfo{year}{2012}\natexlab{}.
\newblock \showarticletitle{Tapping into Knowledge Base for Concept Feedback:
  Leveraging ConceptNet to Improve Search Results for Difficult Queries}. In
  \bibinfo{booktitle}{\emph{Proceedings of the 5th ACM International Conference
  on Web Search and Data Mining (WSDM)}}. ACM, \bibinfo{pages}{403--412}.
\newblock


\bibitem[\protect\citeauthoryear{Krebs, Lenci, and Paperno}{Krebs
  et~al\mbox{.}}{2018}]%
        {krebs2018semeval}
\bibfield{author}{\bibinfo{person}{Alicia Krebs}, \bibinfo{person}{Alessandro
  Lenci}, {and} \bibinfo{person}{Denis Paperno}.}
  \bibinfo{year}{2018}\natexlab{}.
\newblock \showarticletitle{SemEval-2018 Task 10: Capturing Discriminative
  Attributes}. In \bibinfo{booktitle}{\emph{Proceedings of the 12th
  International Workshop on Semantic Evaluation (SemEval)}}.
  \bibinfo{pages}{732--740}.
\newblock


\bibitem[\protect\citeauthoryear{Lao and Cohen}{Lao and Cohen}{2010}]%
        {lao2010relational}
\bibfield{author}{\bibinfo{person}{Ni Lao} {and} \bibinfo{person}{William~W
  Cohen}.} \bibinfo{year}{2010}\natexlab{}.
\newblock \showarticletitle{Relational Retrieval Using a Combination of
  Path-Constrained Random Walks}.
\newblock \bibinfo{journal}{\emph{Machine Learning}} \bibinfo{volume}{81},
  \bibinfo{number}{1} (\bibinfo{year}{2010}), \bibinfo{pages}{53--67}.
\newblock


\bibitem[\protect\citeauthoryear{Lenat}{Lenat}{1995}]%
        {lenat1995cyc}
\bibfield{author}{\bibinfo{person}{Douglas~B Lenat}.}
  \bibinfo{year}{1995}\natexlab{}.
\newblock \showarticletitle{{CYC}: A large-scale investment in knowledge
  infrastructure}.
\newblock \bibinfo{journal}{\emph{Communications of the ACM (CACM)}}
  \bibinfo{volume}{38}, \bibinfo{number}{11} (\bibinfo{year}{1995}),
  \bibinfo{pages}{33--38}.
\newblock


\bibitem[\protect\citeauthoryear{Lin, Liu, Luan, Sun, Rao, and Liu}{Lin
  et~al\mbox{.}}{2015}]%
        {lin2015modeling}
\bibfield{author}{\bibinfo{person}{Yankai Lin}, \bibinfo{person}{Zhiyuan Liu},
  \bibinfo{person}{Huanbo Luan}, \bibinfo{person}{Maosong Sun},
  \bibinfo{person}{Siwei Rao}, {and} \bibinfo{person}{Song Liu}.}
  \bibinfo{year}{2015}\natexlab{}.
\newblock \showarticletitle{Modeling Relation Paths for Representation Learning
  of Knowledge Bases}. In \bibinfo{booktitle}{\emph{Proceedings of the
  Conference Empirical Methods in Natural Language Processing (EMNLP)}}.
  \bibinfo{pages}{705--714}.
\newblock


\bibitem[\protect\citeauthoryear{Liu and Singh}{Liu and Singh}{2004}]%
        {liu2004conceptnet}
\bibfield{author}{\bibinfo{person}{Hugo Liu} {and} \bibinfo{person}{Push
  Singh}.} \bibinfo{year}{2004}\natexlab{}.
\newblock \showarticletitle{{ConceptNet}—a Practical Commonsense Reasoning
  Tool-Kit}.
\newblock \bibinfo{journal}{\emph{BT Technology Journal}} \bibinfo{volume}{22},
  \bibinfo{number}{4} (\bibinfo{year}{2004}), \bibinfo{pages}{211--226}.
\newblock


\bibitem[\protect\citeauthoryear{Mikolov, Sutskever, Chen, Corrado, and
  Dean}{Mikolov et~al\mbox{.}}{2013}]%
        {mikolov2013distributed}
\bibfield{author}{\bibinfo{person}{Tomas Mikolov}, \bibinfo{person}{Ilya
  Sutskever}, \bibinfo{person}{Kai Chen}, \bibinfo{person}{Greg~S Corrado},
  {and} \bibinfo{person}{Jeff Dean}.} \bibinfo{year}{2013}\natexlab{}.
\newblock \showarticletitle{Distributed Representations of Words and Phrases
  and Their Compositionality}. In \bibinfo{booktitle}{\emph{Advances in Neural
  Information Processing Systems (NIPS)}}. \bibinfo{pages}{3111--3119}.
\newblock


\bibitem[\protect\citeauthoryear{Miller, Beckwith, Fellbaum, Gross, and
  Miller}{Miller et~al\mbox{.}}{1990}]%
        {miller1990introduction}
\bibfield{author}{\bibinfo{person}{George~A Miller}, \bibinfo{person}{Richard
  Beckwith}, \bibinfo{person}{Christiane Fellbaum}, \bibinfo{person}{Derek
  Gross}, {and} \bibinfo{person}{Katherine~J Miller}.}
  \bibinfo{year}{1990}\natexlab{}.
\newblock \showarticletitle{Introduction to {WordNet}: An On-line Lexical
  Database}.
\newblock \bibinfo{journal}{\emph{International Journal Lexicography}}
  \bibinfo{volume}{3}, \bibinfo{number}{4} (\bibinfo{year}{1990}),
  \bibinfo{pages}{235--244}.
\newblock


\bibitem[\protect\citeauthoryear{Neelakantan, Roth, and McCallum}{Neelakantan
  et~al\mbox{.}}{2015}]%
        {neelakantan2015compositional}
\bibfield{author}{\bibinfo{person}{Arvind Neelakantan},
  \bibinfo{person}{Benjamin Roth}, {and} \bibinfo{person}{Andrew McCallum}.}
  \bibinfo{year}{2015}\natexlab{}.
\newblock \showarticletitle{Compositional Vector Space Models for Knowledge
  Base Completion}. In \bibinfo{booktitle}{\emph{Proceedings of the 53rd Annual
  Meeting of the Association for Computational Linguistics and the 7th
  International Joint Conference on Natural Language Processing (ACL \&
  IJCNLP)}}. \bibinfo{publisher}{Association for Computational Linguistics},
  \bibinfo{pages}{156--166}.
\newblock


\bibitem[\protect\citeauthoryear{Pennington, Socher, and Manning}{Pennington
  et~al\mbox{.}}{2014}]%
        {pennington2014glove}
\bibfield{author}{\bibinfo{person}{Jeffrey Pennington},
  \bibinfo{person}{Richard Socher}, {and} \bibinfo{person}{Christopher
  Manning}.} \bibinfo{year}{2014}\natexlab{}.
\newblock \showarticletitle{{GloVe}: Global Vectors for Word Representation}.
  In \bibinfo{booktitle}{\emph{Proceedings of the Conference on Empirical
  Methods in Natural Language Processing (EMNLP)}}.
  \bibinfo{pages}{1532--1543}.
\newblock


\bibitem[\protect\citeauthoryear{Radovanovi{\'c}, Nanopoulos, and
  Ivanovi{\'c}}{Radovanovi{\'c} et~al\mbox{.}}{2010}]%
        {radovanovic2010hubs}
\bibfield{author}{\bibinfo{person}{Milo{\v{s}} Radovanovi{\'c}},
  \bibinfo{person}{Alexandros Nanopoulos}, {and} \bibinfo{person}{Mirjana
  Ivanovi{\'c}}.} \bibinfo{year}{2010}\natexlab{}.
\newblock \showarticletitle{Hubs in Space: Popular Nearest Neighbors in
  High-Dimensional Data}.
\newblock \bibinfo{journal}{\emph{Journal of Machine Learning Research (JMLR)}}
   \bibinfo{volume}{11} (\bibinfo{year}{2010}), \bibinfo{pages}{2487--2531}.
\newblock


\bibitem[\protect\citeauthoryear{Singh, Lin, Mueller, Lim, Perkins, and
  Zhu}{Singh et~al\mbox{.}}{2002}]%
        {singh2002open}
\bibfield{author}{\bibinfo{person}{Push Singh}, \bibinfo{person}{Thomas Lin},
  \bibinfo{person}{Erik~T Mueller}, \bibinfo{person}{Grace Lim},
  \bibinfo{person}{Travell Perkins}, {and} \bibinfo{person}{Wan~Li Zhu}.}
  \bibinfo{year}{2002}\natexlab{}.
\newblock \showarticletitle{{Open Mind Common Sense}: Knowledge Acquisition
  from the General Public}. In \bibinfo{booktitle}{\emph{OTM Confederated
  International Conferences ``On the Move to Meaningful Internet Systems''}}.
  Springer, \bibinfo{pages}{1223--1237}.
\newblock


\bibitem[\protect\citeauthoryear{Speer, Chin, and Havasi}{Speer
  et~al\mbox{.}}{2017}]%
        {speer2017conceptnet}
\bibfield{author}{\bibinfo{person}{Robert Speer}, \bibinfo{person}{Joshua
  Chin}, {and} \bibinfo{person}{Catherine Havasi}.}
  \bibinfo{year}{2017}\natexlab{}.
\newblock \showarticletitle{ConceptNet 5.5: An Open Multilingual Graph of
  General Knowledge}. In \bibinfo{booktitle}{\emph{AAAI Conference on
  Artificial Intelligence}}. AAAI, \bibinfo{pages}{4444--4451}.
\newblock


\bibitem[\protect\citeauthoryear{Toutanova, Lin, Yih, Poon, and
  Quirk}{Toutanova et~al\mbox{.}}{2016}]%
        {toutanova2016compositional}
\bibfield{author}{\bibinfo{person}{Kristina Toutanova},
  \bibinfo{person}{Victoria Lin}, \bibinfo{person}{Wen-tau Yih},
  \bibinfo{person}{Hoifung Poon}, {and} \bibinfo{person}{Chris Quirk}.}
  \bibinfo{year}{2016}\natexlab{}.
\newblock \showarticletitle{Compositional Learning of Embeddings for Relation
  Paths in Knowledge Base and Text}. In \bibinfo{booktitle}{\emph{Procedings of
  the 54th Annual Meeting of the Association for Computational Linguistics
  (ACL)}}, Vol.~\bibinfo{volume}{1}. \bibinfo{pages}{1434--1444}.
\newblock


\bibitem[\protect\citeauthoryear{Von~Ahn, Kedia, and Blum}{Von~Ahn
  et~al\mbox{.}}{2006}]%
        {von2006verbosity}
\bibfield{author}{\bibinfo{person}{Luis Von~Ahn}, \bibinfo{person}{Mihir
  Kedia}, {and} \bibinfo{person}{Manuel Blum}.}
  \bibinfo{year}{2006}\natexlab{}.
\newblock \showarticletitle{Verbosity: A Game for Collecting Common-Sense
  Facts}. In \bibinfo{booktitle}{\emph{Proceedings of the SIGCHI Conference on
  Human Factors in Computing Systems (CHI)}}. ACM, \bibinfo{pages}{75--78}.
\newblock


\bibitem[\protect\citeauthoryear{Voorhees}{Voorhees}{2004}]%
        {voorhees2005overview}
\bibfield{author}{\bibinfo{person}{Ellen~M Voorhees}.}
  \bibinfo{year}{2004}\natexlab{}.
\newblock \showarticletitle{Overview of the TREC 2004 Robust Retrieval Track}.
  In \bibinfo{booktitle}{\emph{TREC}}.
\newblock


\bibitem[\protect\citeauthoryear{Vrande{\v{c}}i{\'c} and
  Kr{\"o}tzsch}{Vrande{\v{c}}i{\'c} and Kr{\"o}tzsch}{2014}]%
        {vrandevcic2014wikidata}
\bibfield{author}{\bibinfo{person}{Denny Vrande{\v{c}}i{\'c}} {and}
  \bibinfo{person}{Markus Kr{\"o}tzsch}.} \bibinfo{year}{2014}\natexlab{}.
\newblock \showarticletitle{Wikidata: A Free Collaborative Knowledgebase}.
\newblock \bibinfo{journal}{\emph{Communications of the ACM (CACM)}}
  \bibinfo{volume}{57}, \bibinfo{number}{10} (\bibinfo{year}{2014}),
  \bibinfo{pages}{78--85}.
\newblock


\bibitem[\protect\citeauthoryear{Xiong, Hoang, and Wang}{Xiong
  et~al\mbox{.}}{2017}]%
        {xiong2017deeppath}
\bibfield{author}{\bibinfo{person}{Wenhan Xiong}, \bibinfo{person}{Thien
  Hoang}, {and} \bibinfo{person}{William~Yang Wang}.}
  \bibinfo{year}{2017}\natexlab{}.
\newblock \showarticletitle{DeepPath: A Reinforcement Learning Method for
  Knowledge Graph Reasoning}. In \bibinfo{booktitle}{\emph{Proceedings of the
  2017 Conference on Empirical Methods in Natural Language Processing
  (EMNLP)}}. \bibinfo{pages}{564--573}.
\newblock


\bibitem[\protect\citeauthoryear{Xu, Bai, Bian, Gao, Wang, Liu, and Liu}{Xu
  et~al\mbox{.}}{2014}]%
        {xu2014rc}
\bibfield{author}{\bibinfo{person}{Chang Xu}, \bibinfo{person}{Yalong Bai},
  \bibinfo{person}{Jiang Bian}, \bibinfo{person}{Bin Gao},
  \bibinfo{person}{Gang Wang}, \bibinfo{person}{Xiaoguang Liu}, {and}
  \bibinfo{person}{Tie-Yan Liu}.} \bibinfo{year}{2014}\natexlab{}.
\newblock \showarticletitle{{RC-NET}: A General Framework for Incorporating
  Knowledge into Word Representations}. In \bibinfo{booktitle}{\emph{Procedings
  of the 23rd ACM International Conference Information and Knowledge Management
  (CIKM)}}. ACM, \bibinfo{pages}{1219--1228}.
\newblock


\bibitem[\protect\citeauthoryear{Xu and Croft}{Xu and Croft}{1996}]%
        {Xu:1996:QEU:243199.243202}
\bibfield{author}{\bibinfo{person}{Jinxi Xu} {and} \bibinfo{person}{W.~Bruce
  Croft}.} \bibinfo{year}{1996}\natexlab{}.
\newblock \showarticletitle{Query Expansion Using Local and Global Document
  Analysis}. In \bibinfo{booktitle}{\emph{Proceedings of the 19th Annual
  International ACM SIGIR Conference on Research and Development in Information
  Retrieval}}. \bibinfo{publisher}{ACM}, \bibinfo{address}{New York, NY, USA},
  \bibinfo{pages}{4--11}.
\newblock
\showISBNx{0-89791-792-8}


\bibitem[\protect\citeauthoryear{Zipf}{Zipf}{1935}]%
        {zipf1935psycho}
\bibfield{author}{\bibinfo{person}{George~Kingsley Zipf}.}
  \bibinfo{year}{1935}\natexlab{}.
\newblock \showarticletitle{The Psycho-Biology of Language}.
\newblock  (\bibinfo{year}{1935}).
\newblock


\end{thebibliography}

\end{document}